\documentclass[number,preprint,12pt,sort&compress]{elsarticle}

\newcommand{\NCut}{\mbox{NCut}}
\newcommand{\NAssoc}{\mbox{NAssoc}}
\newcommand{\Curv}{\mbox{Curv}}

\usepackage{cite}
\usepackage{amssymb}
\usepackage{url}
\usepackage{stfloats}
\usepackage{graphicx}
\usepackage[cmex10]{amsmath}
\usepackage{cite}
\usepackage{bm}
\usepackage{amsthm}
\usepackage[linesnumbered, algoruled, longend, boxed]{algorithm2e}
\usepackage{flushend}
\usepackage[font=footnotesize,caption=false]{subfig}
\usepackage{fixltx2e}
\usepackage{stfloats}
\usepackage{url}

\usepackage{multicol}
\usepackage{array}
\usepackage{mdwmath}
\usepackage{mdwtab}

\usepackage{lscape}

\usepackage{hyperref}
\usepackage{chngpage}          
\usepackage{afterpage}
\usepackage{dsfont}

\usepackage{algorithm}
\usepackage{algorithmic}
\usepackage{color}
\definecolor{lightgray}{gray}{0.9}

\usepackage{simplemargins}
	\setallmargins{1in}

\journal{Pattern Recognition}

\begin{document}

\begin{frontmatter}



\title{GANC: Greedy Agglomerative Normalized Cut}

\author{Seyed Salim Tabatabaei\footnote{Corresponding author. Tel.: +1 514 677 0056; fax: +1 514 398 4470; E-mail: \url{seyed.s.tabatabaei@mail.mcgill.ca}.}}
\author{Mark Coates\footnote{E-mail: \url{mark.coates@mcgill.ca}}}
\author{Michael Rabbat\footnote{E-mail: \url{michael.rabbat@mcgill.ca}}}
\address{Department of Electrical and Computer Engineering, McGill University, McConnell Engineering Building, 3480 University Street, Montreal, Quebec, H3A 2A7.}

\begin{abstract}
This paper describes a graph clustering algorithm that aims to minimize the normalized cut criterion and has a model order selection procedure. The performance of the proposed algorithm is comparable to spectral approaches in terms of minimizing normalized cut. However, unlike spectral approaches, the proposed algorithm scales to graphs with millions of nodes and edges. 
The algorithm consists of three components that are processed sequentially: a greedy agglomerative hierarchical clustering
procedure, model order selection, and a local refinement. 

For a graph of $n$ nodes and $O(n)$ edges, the computational complexity of the algorithm is $O(n \log^2 n)$, a major improvement over the $O(n^3)$ complexity of spectral methods. Experiments are performed on real and synthetic networks to demonstrate the scalability of the proposed approach, the effectiveness of the model order selection procedure, and the performance of the proposed algorithm in terms of minimizing the normalized cut metric.

\end{abstract}


\end{frontmatter}


\section{Introduction}
\label{sec:intro}
Clustering or partitioning of nodes in networks or graphs is an
important task that has many applications in a diverse range of
fields. It has been used for many years to study social
networks~\citep{wasserman94} and continues to be employed in the field
of sociology to explore social interactions. More recently it has been
employed in the study of biochemical
networks~\citep{guimera05,loewenstein08}, biological neural networks~\citep{chen08,crum08}, and transport and communication networks.

There are three components to the clustering task: (i) choosing the
number of clusters; (ii) selecting a criterion that measures the merit
of each candidate cluster allocation; and (iii) identifying an
algorithm that searches for the optimal clustering. Some performance
criteria can be used both to select the number of clusters and choose
the clustering, e.g., modularity~\citep{girvan02} or
information-theoretic criteria based on the minimum description
length~\citep{rosvall08}. 

There is no universally-accepted performance criterion and indeed the
most appropriate criterion can vary depending on the application
domain and the goal of the clustering. Modularity, for example,
focuses on network structure and places primary value on direct
connections between nodes. On the other hand, clustering algorithms
based on Markov random walks~\citep{rosvall08,lafon06} also value
indirect connections and network flow. In this paper we select the
normalized cut criterion~\citep{shi00}, which simultaneously encourages
intra-cluster similarity while penalizing inter-cluster similarities.
Methods based on this criterion have been
employed successfully in a wide range of applications~\citep{crum08,lafon06,shi00,fowlkes04,dhillon07,yu10}. 
The normalized cut criterion is related to the conductance of the underlying graph \citep{sinclair89}. Furthermore
an implicit duality between normalized cut and normalized association exists; the former encourages clusters that are less connected to each other and the latter encourages clusters whose nodes are well-connected.

When partitioning a graph into clusters, the cut associated with
cluster $i$ is the sum of the weights of the edges between nodes in
$i$ and nodes in other clusters. Intuitively, minimizing the maximum
cut (or minimizing the average cut) identifies clusters that have the
weakest ties between them. The problem is that the minimum cut will
often be achieved by identifying many very small clusters, which
provides little insight into the underlying structure of the
graph. The normalized cut metric was
introduced by Shi and Malik in~\citep{shi00} to address this
shortcoming. It normalizes each cut by dividing it by the total weight
of the associated cluster. This has the effect of penalizing very
small clusters because they generally have low total weight.

Minimizing normalized cut is an NP-complete problem~\citep{shi00}. Shi and
Malik illustrated that the minimization could be relaxed to form a
generalized eigenvalue system, whose (discretized) solution
corresponds to the minimum normalized cut. This has led to the development of
a spectral clustering; it involves calculating eigenvectors 
of the identified system. In order to identify a partitioning of $n$ nodes to $k$ clusters, some techniques perform
recursive bipartitioning~\citep{shi00}, and thus 
require the repeated identification of two eigenvectors. Other
approaches strive to identify $k$ clusters directly by calculating the
$k$ smallest eigenvectors of the underlying graph Laplacian. One of the
concerns about these methods is computational complexity; even when
employing fast eigendecomposition methods such as the Lanczos
iterative technique, complexity grows rapidly with $n$. 

In this paper, we propose an agglomerative clustering algorithm that
strives to minimize the normalized cut (or equivalently, maximize the
normalized association). It is a fast, scalable algorithm, with almost
linear complexity in the number of nodes for relatively sparse
graphs. Performance evaluation using a range of benchmark and observed
graphs indicates that the algorithm identifies partitionings that have
average normalized association metrics as large as those of the
partitions identified by the spectral clustering
techniques. We also propose a method for identifying important
partitioning scales which can be used to automatically select the number of clusters. To the best of our knowledge, this is the first model order selection method applicable to the normalized association maximization that is scalable. Through our experiments, we will show how effective our model order selection criterion is for both synthetic and real networks. 
 
The rest of the paper is organized as follows. In Section \ref{sec:related} the major approaches for minimization of normalized cut and model order selection are reviewed. The problem formulation of normalized association is re-stated in Section \ref{sec:formulation}. Our proposed algorithm is detailed in Section \ref{sec:ganc}. The proposed algorithm is then compared to the state-of-the-art clustering algorithms in Section \ref{sec:results}. The concluding remarks are provided in Section \ref{sec:conc}. 

\section{Related Work}
\label{sec:related}
The identification of clusters in graphs and networks has received
significant attention. In our review we focus on a representative set of algorithms that minimize the normalized cut with or without spectral decomposition. We also review the existing methods for selecting the number of clusters. 

\subsection{Optimization of Normalized Cut}
Spectral partitioning of
graphs was first proposed by Donath and Hoffman in the
1970's~\citep{donath72}. Interest in the techniques was renewed in the
1990's when Pothen et al.\ described an algorithm for bi-partitioning
using the Fiedler vector~\citep{pothen90}. Hendrickson et al.\ and
Karypis et al.\ contributed with multilevel algorithms for more
efficient spectral partitioning~\citep{Hendrickson95,Karypis99}.

The normalized cut metric was introduced by Shi and Malik
in~\citep{shi00}. They demonstrated how the bipartitioning task, with
the objective of minimizing the normalized cut, could be relaxed to
construct a generalized eigenvalue problem, and was thus related to
spectral partitioning. The eigenvector corresponding to the
second-smallest eigenvalue of the graph Laplacian identifies the bipartitioning (the real
values in the eigenvector must be mapped to two discrete values for
partitioning). This established the connection between minimization of
normalized cut and spectral partitioning.  Shi and Malik proposed a recursive bipartitioning scheme in order to partition a graph into $k$ clusters. In~\citep{meila01}, Meila and Shi proposed an algorithm that
calculates $k$ eigenvectors (thereby associating $k$ real values with
each node in the graph) and then uses a clustering algorithm, such as
$k$-means, to do the partitioning in $\Re^k$. Ng et al.\ observed
in~\citep{ng02} that the algorithm in~\citep{meila01} is susceptible to
failure when there is substantial variation in the degree of
connectivity between clusters. They proposed an alternative algorithm
that uses a different normalization in both the eigenvalue problem and
the construction of feature vectors prior to the application of
$k$-means. 

All of these algorithms involve a computationally expensive
eigendecomposition. To address the computational difficulties for
large graphs, Fowlkes et al.\ described a procedure that uses the
Nystr{\"o}m method to reduce the complexity of the eigenvalue
problem~\citep{fowlkes04}. This does significantly reduce the
computational overhead, but it is not enough
to make the eigendecomposition algorithms scalable to large
graphs. Yan et al.\ recently described an algorithm for fast
approximate spectral clustering~\citep{yan09}, but the focus is not on
clustering for graphs (rather it addresses real-valued feature
vectors). 

Dhillon et al.\ introduced a much faster algorithm for
minimizing normalized cut in~\citep{dhillon07}; the
graph is first greedily coarsened, then the coarsened graph is
partitioned using a region growing procedure~\citep{Karypis99} and finally weighted kernel
$k$-means clustering is applied to each partition to refine the
clustering. 

Other methods also exist that strive to optimize related cost functions; for example Sharon et al.\ proposed a scalable hierarchical clustering using {\em ratio association} (the normalization term is the number of nodes in clusters) \citep{sharon00}. Other examples include the work by Ding et al.\ \citep{ding01}, Sharon et al.\ \citep{sharon01}, Akselrod-Ballin et al.\ \citep{akselrodballin06}, and Corso et al.\ \citep{corso08} that use the sum of internal weights of clusters as the normalization term.

\subsection{Model Order Selection}
The problem of selecting the number of clusters is almost as old as clustering itself. In the following discussion, $k^*$ denotes the number of clusters that a given approach identifies as the true number of clusters. 

Numerous authors have used some notion of quality of clustering, usually based on some definition of the inter- and intra-cluster distances,  to identify the number of clusters. Let $F(k)$ denote this notion for a given partitioning to $k$ clusters. $F(k)$ is then examined for various values of $k$. Partitionings for different values of $k$ can be obtained from a hierarchical clustering, or alternatively, by several flat partitionings.  

If $F$ is not a monotonic function of $k$, the approach is usually to take $k^* = \arg\max_kF(k)$ or $k^* = \arg\min_kF(k)$ (depending on the definition of $F$). The majority of the methods discussed in \citep{milligan85} and \citep{halkidi01} are of this type; Calinski and Harabasz \citep{calinski74}, C-index \citep{hubert76}, Baker and Hubert \citep{baker75}, and cubic clustering criterion \citep{sarle83} are among the most effective ones \citep{milligan85}. Tibshirani et al.\ \citep{tibshirani01estimating} define $F(k)$ as the {\em gap}, that is the difference of the average pairwise distance of the data points of the clustering at $k^{\text{th}}$ level and the expected value of the same measure of some reference model. This is similar to modularity \citep{girvan02} (discussed later) in the sense that it compares the clustering results to a reference model. 
None of these methods are directly applicable to graph clustering algorithms; the calculations of the defined metrics require pairwise distances which are not immediately available from a graph representation. Possible distance metrics include the shortest path \citep{edachery99} or the diffusion distance \citep{lafon06}; however the shortest path is very sensitive to noise and the calculation of the diffusion distance requires eigendecomposition. 

In the case where $F$ is a monotonic function of $k$, the only extremal values of $F(k)$ correspond to trivial values of $k$. Hence, the value of $F(k)$ corresponding to two or more choices of $k$ are examined to quantify the significance of a given level of a hierarchical clustering~\citep{gnanadesikan77,krzanowski88,pedersen06}. In order to assess the clustering at level $k$ of the hierarchy, Gnanadesikan et al.\ \citep{gnanadesikan77} propose the fraction $F(k)/F(k-1)$, Arnold \citep{arnold79} use the value of $F(k)-F(k-1)$, Krzanowski and Lai \citep{krzanowski88} employ the fraction $|F(k) - F(k-1)|/ |F(k+1) - F(k)|$, and Pederson and Kulkarni \citep{pedersen06} suggest using the fraction $2 \times F(k) / \left( F(k-1) + F(k+1) \right)$. The latter two approaches are the most similar to what we propose in Section \ref{sec:ganc} in the sense that they use the preceding and succeeding levels of the hierarchy to obtain the significance of the clustering at level $k$. However the approaches in \citep{krzanowski88,pedersen06} are potentially susceptible to noise due to the division; small perturbations in the weights could lead to dramatic changes in the selected model order.


Other approaches to selecting the number of clusters are based on the adoption of (semi-)parametric models for the structure of the graph. This allows the application of model selection techniques based on concepts such as the Bayesian InformationCriterion (BIC) \citep{schwarz1978}, and Akaike Information Criterion \citep{akaike74}. Despite the strong theoretical support for these methods, the adoption of a parametric model for the graph structure is undesirable. The models are often overly restrictive and do not adequately capture the properties of many real-life networks. An example is the requirement in \citep{wolfe70,pelleg00} that the input data are normally distributed (after projection of  the graph into a real space). The more general methods based on mixture models do not scale well to very large graphs; even the recent approaches have only been applied to graphs with a few thousand nodes \citep{mariadassou10,latoucherep08,latouche10}. 

Some heuristics are based on the sizes of the clusters that are merged at different levels of the clustering hierarchy \citep{gdalyahu99,harel01}. The authors in \citep{harel01} suggest that when two clusters with large number of nodes are merged, a significant amount of detail is lost; hence such an instance is potentially where a hierarchical clustering algorithm should stop. The authors of \citep{gdalyahu99} propose a similar approach. The main drawback of these approaches is that only the granularity of the clusters are taken into account and the number and the weights of the edges are simply ignored. 

A well-known and effective method of selecting the number of clusters is to examine the eigenvalues of the Laplacian of the graph that is to be clustered \citep{ng02, azran06, vonLuxburg07}. For a graph with isolated connected components the multiplicity of the eigenvalue zero is equal to the number of clusters. Any other graph with well-separated clusters can be considered as a perturbation of this ideal case. Matrix perturbation theory states that the stability of the eigenvectors of a matrix is proportional to the eigengap (the difference between two successive eigenvalues). Von Luxburg \citep{vonLuxburg07} suggests using $k^* = \arg\max_k \left( \lambda_{k}-\lambda_{k+1} \right)$. A large eigengap at $k^*$ is the case in which spectral algorithms using the Laplacian perform most successfully \citep{ng02}. A more robust criterion is proposed by Zelnik-Manor and Perona \citep{ZelnikManor04} that uses eigenvectors instead. Despite the solid theoretical support behind these approaches, the requirement of eigendecomposition makes them impractical when large graphs are being clustered. 

There have been attempts to identify the number of clusters using stability analysis. ``Stability'' has been defined differently by different authors, but they all based on the same intuition; with respect to some algorithm, a stable clustering is one that behaves relatively consistently in the presence of some controlled perturbation. The perturbation could be in terms of noise \citep{ben-hur02}, sampling subsets from the input \citep{lange04,ben06}, or random projection of a high dimensional data into a lower dimensional space \citep{smolkin03}. These approaches require several runs of clustering for every value of $k$ which makes them computationally expensive. Furthermore, Ben-David et al.\ \citep{ben06} warn against using stability analysis in this context, and they suggest that this family of model order selection techniques is not suitable for selecting the number of clusters in general. The intuition behind their work is that when an underlying objective function, $F(k)$, has several local optima of relatively similar values, a clustering algorithm might be trapped by any of them, even when $k$ is the true number of clusters. The difference in clustering solutions is interpreted as instability and the model order rejected, but the behavior is caused by the imperfect nature of the clustering algorithm. 


Two more recently proposed methods to select the number of clusters are based on optimization of the quality metrics of modularity \citep{girvan02} and description length \citep{rosvall08}. Both methods simultaneously address both clustering and the selection of the number of clusters; they strive to optimize the objective function over all possible partitionings.
These methods are discussed in more detail in Section \ref{sec:results}.

\section{Problem Formulation}
\label{sec:formulation}
Let $G = (V,E,w)$ be a weighted graph having $n = |V|$ nodes and $m
= |E|$ edges\footnote{In this work, we assume we are given the graph
  on which we wish to perform clustering.  We do not address the
  problem of forming a graph from data, which arises when one applies
  graph clustering methods to general data sets; see, e.g.,
  \citep{vonLuxburg07,daitch09,maier09}.}.  We assume that edge weights
$w(u,v) = w(v,u) \ge 0$ are non-negative and symmetric, that $w(u,v) =
0$ if $(u,v) \notin E$, and that for $(u,v) \in E$, the weight
$w(u,v) \geq 0$ is indicative of the similarity between nodes $u$ and $v$;
that is, the larger the weight, the more similar the nodes.  We
also allow for self-weights, $w(u,u) \ge 0$.  

For a fixed number of clusters, $k$, we measure the quality of the
partition via the normalized cut metric~\citep{shi00}, defined as
follows.  For a node $u$, denote the degree of $u$ by $d(u) =
\sum_{v \in V} w(u,v)$, and for a subset $U \subset V$ of nodes,
let $d(U) = \sum_{u \in U} d(u)$ denote the cumulative degree of the
subset.  Similarly, for two disjoint subsets of nodes, $V_1$ and
$V_2$, let $w(V_1, V_2) = \sum_{u \in V_1} \sum_{v \in V_2} w(u,v)$
denote the sum of the weights of edges with one end in $V_1$ and the
other end in $V_2$. Let $\mathcal{C}_k$ denote the partition of nodes to $k$ clusters, $\mathcal{C}_k=\{ C_1,\dots,C_k \}$ where $C_i$ is the subset of nodes affiliated to cluster $i$. The \emph{normalized cut} metric is defined as
\begin{equation}
\NCut(\mathcal{C}_k) = \sum_{i=1}^k \frac{w(C_i, V \setminus C_i)}{d(C_i)}.
\end{equation}
Minimizing the normalized cut can be interpreted as minimizing the
similarity of nodes in different clusters, relative to the degree
of each cluster.  Alternatively, maximizing the intra-cluster
similarity can be achieved by maximizing the \emph{normalized
  association}, defined as
\begin{equation}
\NAssoc(\mathcal{C}_k) = \sum_{i=1}^k \frac{w(C_i, C_i)}{d(C_i)}. \label{eq:nassoc}
\end{equation}
Moreover, maximizing normalized association is equivalent to minimizing
normalized cut since $w(C_i, C_i) + w(C_i, V \setminus C_i) = d(C_i)$,
and hence $\NAssoc(\mathcal{C}_k) = k - \NCut(\mathcal{C}_k)$.  For the sake of
readability, we adopt the maximization of normalized association as our goal in developing a clustering procedure.

\section{GANC: Greedy Agglomerative Normalized Cut} \label{sec:algorithm}
\label{sec:ganc}
In this section, we describe a greedy algorithm for building an agglomerative
clustering on a graph.  Although we do not make guarantees about its
accuracy, the algorithm is fast on sparse graphs and yields excellent
performance on a variety of examples, as illustrated in
Section~\ref{sec:results}.  Moreover, GANC has a model order selection criterion and does not have to be provided 
with the number of clusters a priori. 
The algorithm consists of three steps: agglomerative clustering, model order selection, and refinement.

\subsection{Agglomerative Clustering}
We propose greedy maximization of normalized association via an agglomerative hierarchical clustering. In the following discussion, note that the number of clusters at stage $k$ of the hierarchy is $k$. At stage $k$ of the hierarchy, using the given partition, $\mathcal{C}_k$, two clusters are merged to form a new partition, $\mathcal{C}_{k-1}$. The two clusters to be merged are chosen greedily so that normalized association of stage $k-1$ is maximized.\footnote{We note that a
  similar greedy merging algorithm is alluded to by Shi and
  Malik in~\citep{shi00}, however they also suggest first projecting each
  node into $\Re^k$ using the first $k$ eigenvectors of the graph
  Laplacian, and running an algorithm such as $k$-means to obtain an
  initial clustering.  Our approach requires no eigendecomposition and
  performs no such initialization.  We also note that, although Shi
  and Malik mention having experimented with greedy merging, results
  for this approach have not been reported in the literature.}

Initially, $\mathcal{C}_n$ is a function that maps nodes to unique clusters, $1,\dots,n$. The degree of every node $u$, $d(u)$, is calculated. Furthermore, for every edge, $(u,v) \in E$, the improvement in normalized association by its contraction to obtain $\mathcal{C}_{n-1}$ is stored in $\Delta(u,v)=2w(u,v)/\left(d(u)+d(v) \right)$, so that a matrix of the improvements is constructed. Here we assume that initially there are no self-loops, but if there are, the value of improvements are computed by (\ref{eq:greedy_improvement}) presented shortly. Throughout the agglomerative clustering procedure, $d(\cdot)$, $w(\cdot,\cdot)$, and $\Delta(\cdot,\cdot)$ are updated at each stage as described below.

At each iteration, the pair $(u^*,v^*)=\text{argmax}_{(u,v)} \Delta(u,v)$ is selected for merging to create a larger cluster, $uv^*$. The degree is computed for the newly constructed cluster, $d(uv^*)=d(u^*)+d(v^*)$. The weights and the improvement matrices are updated by removing the rows and columns corresponding to $u^*$ and $v^*$, and inserting a new row and column corresponding to $uv^*$ (rows and columns are not removed or added in our implementation, but this is the practical effect). The weight matrix is updated as follows:
\begin{align}
  w(uv^*,x)=&w(x,uv^*) = w(u^*,x)+w(v^*,x), \label{eq:greedy_update_1}
\end{align}
and the improvement matrix update is:
\begin{align}
\Delta(uv^*,x) = \frac{w(uv^*,uv^*)+w(x,x)+2w(uv^*,x)}{d(uv^*)+d(x)} -\frac{w(uv^*,uv^*)}{d(uv^*)} - \frac{w(x,x)}{d(x)}, \label{eq:greedy_improvement}
\end{align}
for all the clusters, $x$, adjacent to either $u^*$ or $v^*$. The self-weights are also calculated by:
\begin{align}
  w(uv^*,uv^*)=w(u^*,u^*) + w(v^*,v^*) + 2w(u^*,v^*). \label{eq:greedy_update_2}
\end{align}

For all pairs of nodes $(u,v)$ not adjacent to $u^*$ and $v^*$, the weights $w(u,v)$ and improvements $\Delta(u,v)$ remain unchanged. The above sequence of steps is repeated $n-1$ times to form the clustering hierarchy. 

\subsection{Model Order Selection}
Many of the clustering algorithms require the number of clusters to be provided to the algorithm a priori. However such information is often not available in practical situations making the decision about the number of clusters an issue in itself.  

It is worth noting that the stage number ($k$) that maximizes $\NAssoc(\mathcal{C}_k)$ does not necessarily correspond to a meaningful number of clusters.\footnote{To see this, consider an unweighted graph that consists of two isolated chains, each having 4 nodes. The true number of clusters is trivially 2 resulting in $\NAssoc(\mathcal{C}_2)=2$; however if one groups the adjacent nodes to obtain 4 groups of node pairs, the resulting value of normalized association would be $\NAssoc(\mathcal{C}_4)=2.66$.} Here we propose a simple but effective approach to model order selection. 
 Let $\mathcal{C}_k^* = \text{argmax}_{\mathcal{C}_k} \NAssoc(\mathcal{C}_k)$ denote the partition that maximizes normalized assocation over all partitions of $V$ into $k$ clusters. To carry out model order selection, we examine the \textit{curvature}\footnote{The reason we call this metric the curvature is its similarity to the central approximation of the second order derivative which is defined for a continuous function, $f(\cdot)$ as $f''(x) \approx {\partial^2_h[f](x)}/{h^2} = \left[{f(x+h)-2f(x)+f(x-h)}\right]/{h^2}$. By substituting $h=1$, we get the negative of our curvature equation (the negation is to have positive peaks).} of $\NAssoc(\mathcal{C}_k^*)$ which we define as
\begin{align}
  \Curv(k) = &\left( \NAssoc(\mathcal{C}_k^*) - \NAssoc(\mathcal{C}_{k-1}^*)  \right) -\left( \NAssoc(\mathcal{C}_{k+1}^*) - \NAssoc(\mathcal{C}_k^*) \right) \notag \\
	  = & 2 \NAssoc(\mathcal{C}_k^*) - \NAssoc(\mathcal{C}_{k-1}^*) - \NAssoc(\mathcal{C}_{k+1}^*) \label{eq:curv_2} . 
\end{align}
The first term of the above addition is the improvement of normalized association moving from stage $k$ to $k-1$ of the hierarchy and the second term is the improvement moving from stage $k+1$ to $k$. 

The function $\Curv(k)$ captures the notion that a particular number of clusters, $k$, identifies meaningful structural similarities embodied in the graph if it provides a normalized association which is significantly larger than the best partition with one additional cluster $(k+1)$ and little can be gained by reducing the number of clusters to $k-1$. 

In practice we do not have access to the optimal partitions, so we cannot evaluate the exact value of the curvature function. Instead, we approximate the curvature by using the normalized association values for the partitions returned by the agglomerative step of our algorithm. 

Note that the model order selection step of the algorithm could be used as a model order selection step for any other algorithm that maximizes the normalized association and generates a clustering hierarchy. Furthermore, this step of the algorithm can be considered optional if there is a prior knowledge about the true number of clusters. 

\subsection{Refinement}
\label{subsec:refinement}
Greedy algorithms can get trapped by local optima. This is also the case with the agglomerative step of our algorithm, especially when the clusters are not clearly separated. After selecting the number of clusters either using the model order selection rule described previously or prior knowledge, a refinement step is invoked in order to improve the initial clustering results. The nodes are moved across the clusters to further improve the value of normalized association. A similar approach is taken in \citep{blondel08}, but groups of nodes are moved from a cluster to another instead of individual nodes. If we try all possible moves, we end up performing an exhaustive search of dividing $n$ nodes into $k$ clusters. This defeats the purpose of developing the fast agglomerative clustering step. Instead, we look only at the boundary nodes, i.e., the nodes that have at least one neighbor in another cluster. 

Initially the set of all the boundary nodes is identified. We use an $n \times k$ matrix to keep track of the neighborhood information of the boundary nodes. If $B$ denotes this matrix, and $I$ denotes an $n$-dimensional vector:
\begin{align}
  B(u,i) = \left\{ \begin{array}{ll} \sum_{v \in C_i} w(u,v) & \mbox{if } u \not \in C_i  \\
				      0 & \mbox{otherwise}  \end{array} \right.,~
      I(u)=\sum_{v \in C_i} w(u,v) ~\mbox{if }u \in C_i.
\label{eq:B_and_I}
\end{align}

For each of the boundary nodes, improvements in normalized association by moving from their current cluster to each of their neighbors are calculated. This improvement for node $u$ moving from cluster $i$ to cluster $j$ is
\begin{align}
\delta(u,i,j) = &\frac{w(C_i,C_i) - 2I(u)}{d(C_i)-d(u)} + \frac{w(C_j,C_j) + 2B(u,j)}{d(C_j)+d(u)} \notag \\
		  &- \left( \frac{w(C_i,C_i)}{d(C_i)} + \frac{w(C_j,C_j)}{d(C_j)} \right) \label{eq:refinement_delta}.
\end{align}

If no move leads to improvement, the node stays in the cluster to which it belongs. If one or more moves result in improvement, then the node is moved to the cluster that results in maximum improvement. 
If $u$ is moved from $C_i$ to $C_j$, the corresponding cluster degrees and associations are updated:
\begin{align}
  d(C_i)^{\text{new}} &= d(C_i)^{\text{old}} - d(u) \label{eq:refinement_update_1}\\
  d(C_j)^{\text{new}} &= d(C_j)^{\text{old}} + d(u) \label{eq:refinement_update_2}\\
  w(C_i,C_i)^{\text{new}} &= w(C_i,C_i)^{\text{old}} - 2I(u,i)^{\text{old}} \label{eq:refinement_update_3}\\
  w(C_j,C_j)^{\text{new}} &= w(C_j,C_j)^{\text{old}} + 2B(u,j)^{\text{old}} \label{eq:refinement_update_4}, 
\end{align}
and for all the neighboring nodes of $u$ denoted by $v$, entries of $B$ and $I$ are updated as follows:
\begin{align}
  \left.
  \begin{aligned}
        I(v)^{\text{new}} &= I(v)^{\text{old}} + w(u,v) \\ B(v,i)^{\text{new}} &= B(v,i)^{\text{old}} - w(u,v)  
  \end{aligned}
  \right\}
  \label{eq:refinement_update_5}
  &~~~\text{if $v \in C_j$},
\\
  \left.
  \begin{aligned}
  I(v)^{\text{new}} &= I(v)^{\text{old}} - w(u,v) \\ B(v,j)^{\text{new}} &= B(v,j)^{\text{old}} + w(u,v) 
  \end{aligned}
  \right\}
  \label{eq:refinement_update_6}
  &~~~\text{if $v \in C_i$},
\\
  \left.
  \begin{aligned}
  B(v,i)^{\text{new}} &= B(v,i)^{\text{old}} - w(u,v) \\ B(v,j)^{\text{new}} &= B(v,j)^{\text{old}} + w(u,v) 
  \end{aligned}
  \right\}
  \label{eq:refinement_update_7}
  &~~~\text{if $v \notin C_i$ and $v \notin C_j$},
\end{align}
and finally
\begin{align}
  &B(u,i)^{\text{new}} = I(u)^{\text{old}}  \label{eq:refinement_update_8}, \\ 
  &I(u)^{\text{new}} = B(u,j)^{\text{old}} \label{eq:refinement_update_9}, \\
  &B(u,j)^{\text{new}} = 0  \label{eq:refinement_update_10}. 
\end{align}

While performing the updates (\ref{eq:refinement_update_1}) to (\ref{eq:refinement_update_10}), the set of boundary nodes is also updated. Whenever a node is moved from cluster $i$ to cluster $j$, its neighbors in cluster $i$ are added to the set of boundary nodes. For some node $v$ neighboring $u$, moving $u$ might result in $B(v,i)^{\text{new}}$ becoming zero in (\ref{eq:refinement_update_7}) for all $i$; i.e., some nodes might be removed from the set of boundary nodes. 

 One pass through all the boundary nodes is considered a single refinement \textit{iteration}. When an iteration causes no positive improvement, the refinement procedure is stopped. Alternatively, one could specify the maximum number of refinement iterations. 
Note that although we prohibit the refinement step from emptying any cluster, we have not observed such an attempt in our experiments. 

\begin{algorithm}[!th]
\colorbox{lightgray}{GREEDY AGGLOMERATION}
\begin{algorithmic}[1]
  \STATE Build the initial $\Delta$ matrix, and $\NAssoc(\mathcal{C}_n) \gets \sum_{u\in V} w(u,u)/d(u)$
  \FOR{$k=n-1$ to 1} 
    \STATE $(i^*,j^*) \gets \arg\max_{i,j} \Delta(i,j)$
    \STATE $\NAssoc(\mathcal{C}_k) \gets \NAssoc(\mathcal{C}_{k+1}) + \Delta(i^*,j^*)$
    \STATE Update rows and columns of $\Delta$, the weights, and the degrees corresponding to clusters $i^*$ and $j^*$  (\ref{eq:greedy_update_1} - \ref{eq:greedy_update_2})
  \ENDFOR
\end{algorithmic}
\colorbox{lightgray}{MODEL ORDER SELECTION}
\begin{algorithmic}[1]
  \FOR{$k=n-1$ to 2}
    \STATE Calculate $\Curv(k)$ (\ref{eq:curv_2})
  \ENDFOR
  \STATE $k^* \gets \arg\max_k \Curv(k)$ or provided by the user
\end{algorithmic}
\colorbox{lightgray}{REFINEMENT}
\begin{algorithmic}[1]
  \STATE Obtain flat partitioning for level $k^*$
  \STATE Build the initial $B$ and $I$ (\ref{eq:B_and_I}), and $\text{\em refine} \gets$ \TRUE
  \WHILE{$\text{\em refine}$}
    \STATE $\bar{\delta} \gets 0$
    \FOR {$u=1$ to $n$}
      \IF {$u \in C_i$ is on the boundary and $\max_j \delta(u,i,j) > 0$}
	\STATE Move $u$ from $C_i$ to $C_{j^*}$, where $j^*=\arg\max_j\delta(u,i,j)$
	\STATE Update $B$, $I$, the weights, and the degrees (\ref{eq:refinement_update_1} - \ref{eq:refinement_update_10})
	\STATE $\bar{\delta} \gets \bar{\delta} + \delta(u,i,j^*)$
      \ENDIF
    \ENDFOR
    \IF {$\bar{\delta} = 0$} 
      \STATE $\text{\em refine} \gets $ \FALSE 
    \ENDIF
  \ENDWHILE
\end{algorithmic}
\caption{GANC: Greedy Agglomerative Normalized Cut}
\label{alg:GANC}
\end{algorithm}

\subsection{Implementation and Computational Complexity of GANC}
\label{subsec:implementation_and_computation}
We take a similar approach to \citep{clauset04} when implementing the agglomerative clustering procedure of GANC. Max-heaps and balanced binary trees are used to store the rows of the $\Delta$ matrix and the adjacency matrix. A separate heap is also used to store the maximum of each row of $\Delta$. This leads to the complexity of $O(mh\log(n))$ for the agglomerative clustering procedure, where $h$ is the height of the generated dendrogram and $m$ is the number of edges with non-zero weight (see \citep{clauset04} for details). 

The model order selection step of the algorithm requires $O(n)$ computations. The computational requirements are much less than those of the methods analyzing the eigenvalue fall-off which require eigendecomposition (e.g., \citep{ZelnikManor04} and \citep{azran06}). Performing an eigendecomposition to study the eigenvalue fall-off for all of the eigenvalues requires $O(n^3)$ operations. 

To implement the refinement step, we use the map data structure to store and update $B$. Every row of $B$ is stored in a map (C++ STL implementation of the map data structure is used \citep{plauger2000cpp}). Updates (\ref{eq:refinement_update_1}) to (\ref{eq:refinement_update_4}) are performed in constant time. To access each map entry of a row of $B$ to insert, update, or delete, no more than $O(\log k)$ operations are required, because the maximum number of clusters connected to a boundary node is strictly less than $k$. Updates (\ref{eq:refinement_update_5}), (\ref{eq:refinement_update_6}), and (\ref{eq:refinement_update_7}) each take no more than $O(\log k)$ operations and are repeated for all neighbors of a node that is moved. We have observed through experiments that nodes with larger degrees do not tend to be on the boundaries and hence the average number of neighbors to be updated is practically smaller than the average node degree.  (\ref{eq:refinement_update_8}), (\ref{eq:refinement_update_10}), and any insertion to or deletion from the set of boundary nodes are performed in $O(\log k)$ time. (\ref{eq:refinement_update_9}) is trivial and is performed in a constant time. Not more than $n$ nodes can be moved in a given iteration. When node $i$ is moved, $O(d_i \log k)$ operations are required, where $d_i$ is the degree of node $i$. Summing over all boundary nodes, the complexity is $O(m\log k)$ because $\sum_{i=1}^n d_i = 2m$. Hence the number of operations in a single iteration is at most $O(m\log k)$. 
Throughout our experiments we have observed very small number of refinement iterations even when the refinement is applied to the networks with millions of nodes and edges. Furthermore the algorithm is capable of limiting the number of iterations. Hence the complexity of the average number of operations of the whole refinement step does not exceed $O(m \log k)$. 

The total computational complexity of GANC is dominated by the agglomerative clustering procedure. In practice, the height of the resulting dendrogram is often $O(\log n)$. Also, the graphs that are studied are usually sparse and hence $m=O(n)$. Therefore the complexity of the agglomerative clustering step of our algorithm is $O(n \log^2 n)$ for many real-life graphs. 

The worst case runtime happens when the graph is complete and the dendrogram is totally unbalanced. There are $n-1$ agglomerations to construct the whole hierarchy each of which requires at most $O(n\log n)$ operations. Hence the worst case complexity for the agglomerative clustering step of the algorithm is $O(n^2 \log n)$ which is not expected to occur in practical situations.

The memory requirement of GANC is $O(m)$ which is equivalent to $O(n)$ for sparse graphs. In summary, in many practical cases (relatively sparse graphs and balanced dendrograms), the computational complexity is $O(n\log^2 n)$ and the memory requirement is $O(n)$. GANC can be downloaded from {\url{http://www.ece.mcgill.ca/~coates/software/}}.

\section{Experimental Results} \label{sec:results}

In this section we provide a comparison of the performance and the runtime of our proposed algorithm GANC with a selection of the state-of-the-art graph clustering algorithms from the literature whose implementations were downloaded from the corresponding authors' websites. Experiments were performed on an Intel 3.0 GHz Core 2 Quad CPU with 8 GB of RAM and Ubuntu 9.10 operating system. 

\subsection{Comparing Algorithms}
We compare to the following algorithms: Shi and Malik (recursive NCut)~\citep{shi00}; Meila and Shi (k-way NCut)~\citep{meila01}; Ng, Jordan, and Weiss~\citep{ng02}; Dhillon, Gaun, and Kulis~\citep{dhillon07}; Rosvall and Bergstrom~\citep{rosvall08}; and Blondel et al.\ \citep{blondel08}. The first four algorithms focus on maximizing $\NAssoc$. Despite not addressing our criterion of interest directly, the other algorithms are included because they are scalable and represent the state-of-the-art in graph clustering.



The discussed clustering algorithms (\citep{shi00},\citep{meila01},\citep{ng02}) are not scalable as they include eigendecomposition. Dhillon, Guan, and Kulis proposed an algorithm that strives to maximize normalized association without requiring any eigendecomposition \citep{dhillon07}. Similar to the other algorithms  that address the maximization of normalized association, the Dhillon-Guan-Kulis algorithm requires the user to provide the number of clusters. After specifying the number of clusters, three steps are performed: a coarsening phase, a base clustering step using region growing, and finally a refinement step using weighted kernel k-means with appropriate choices of the kernel and the node weights to maximize the normalized association. 

Minimization of the criterion proposed by Rosvall and Bergstrom \citep{rosvall08} results in a coarse-grained representation of the information flow through the network. Their proposed clustering objective aims to compress the underlying random walk on a graph without losing important network structures. To facilitate this, a coding is designed as follows: every cluster is assigned to a unique code; in a given cluster, every node is also assigned to a unique code. However two nodes in different clusters are allowed to be assigned to the same code. The algorithm then strives to minimize the average description length of a single step of a random walk. 

The algorithm by Blondel et al.\ targets maximizing modularity \citep{blondel08}. Modularity ($Q$) is defined as the total fraction of intra-cluster sum-weight minus the expected fraction if the edges (and weights) were distributed randomly while the node degrees were preserved:
\begin{align}
  Q = \sum_{i=1}^{k} \left( \frac{w(C_i,C_i)}{M} - \frac{d(C_i)^2}{M^2} \right), \label{eq:modularity}
\end{align}
where $M=\sum_i\sum_jw(i,j)$. Modularity provides a valuable metric of the connectedness of clusters, but a number of authors have demonstrated that it suffers from a resolution limit when used to select the number of clusters \citep{good09}. The partitioning that maximizes modularity will generally not isolate clusters if the number of edges in the cluster is a small fraction of the total number of edges in the graph. 
The reason behind the resolution limit of modularity is the second term in the summation of (\ref{eq:modularity}); when the network gets larger, $M$ increases monotonically. However the cluster degrees are not necessarily a function of the network size and are often bounded, regardless of the number of nodes \citep{lekovec2008natural}. This results in ${d(C_i)^2}/{M^2} \ll 1$ and hence the value of modularity becomes dependent only on ${w(C_i)}/{M}$. By normalizing the summation by $d(C_i)$ as suggested in~\citep{yu10}, the resolution limit phenomena is resolved; but we have
\begin{align}
   Q_{\text{normalized}} &= \sum_{i=1}^{k} \frac{1}{d(C_i)}\left( \frac{w(C_i,C_i)}{M} - \frac{d(C_i)^2}{M^2} \right) = \frac{1}{M} \left[ \NAssoc(\mathcal{C}_k) - 1 \right].
\end{align}
i.e., maximization of $Q_{\text{normalized}}$ and normalized association are equivalent. 


\subsection{Synthetic Graphs}
We first analyze the performance on synthetic graphs, for which the true clustering behavior is known. We use benchmark graphs developed by Lancichinetti, Fortunato, and Radicchi \citep{lancichinetti09} (LFR graphs). These random graphs are designed based on the planted partition model \citep{condon2001algorithms}. Each node is assigned to one of $k$ clusters. When edges are added to the graph, the probability that the edge is between nodes in the same cluster is $1-\mu$ and the probability that the edge joins nodes from different clusters is $\mu$. The LFR benchmarks have heterogeneous cluster sizes with user-specified lower bound and upper bound, $c_{min}$ and $c_{max}$, respectively. Furthermore the node degrees are upper-bounded by $d_{max}$ and the average node degree is denoted by $d_{avg}$. As $\mu$ decreases, edges are increasingly likely to be intra-cluster, making the partitioning task easier. 

The Meila-Shi algorithm performs at least as well as the other spectral clustering algorithms (Ng-Jordan-Weiss \citep{ng02} and Shi-Malik \citep{shi00}) and hence we only display the Meila-Shi results. The local search of Dhillon-Guan-Kulis is also excluded often, because it results in negligible improvement while it introduces a very large computational overhead. 

\begin{figure}[ht]
\centering 
\subfloat[$n=5,000$] {\hspace*{-.4in} \includegraphics[width=.55\linewidth]{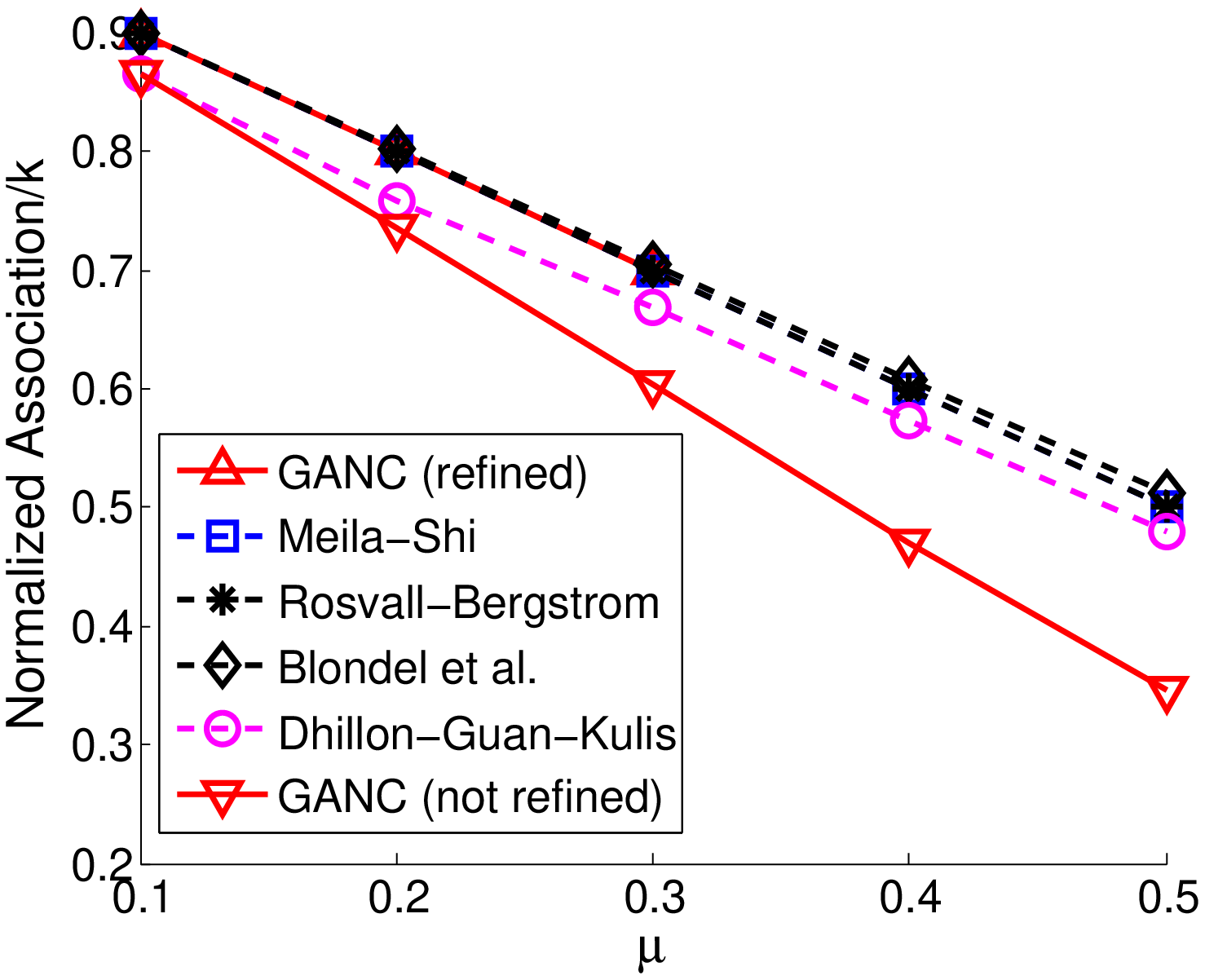} \label{fig:LFR:fig:LFR_5000_nassoc}}
\subfloat[$n=5,000$] {\hspace*{-.1in} \includegraphics[width=.55\linewidth]{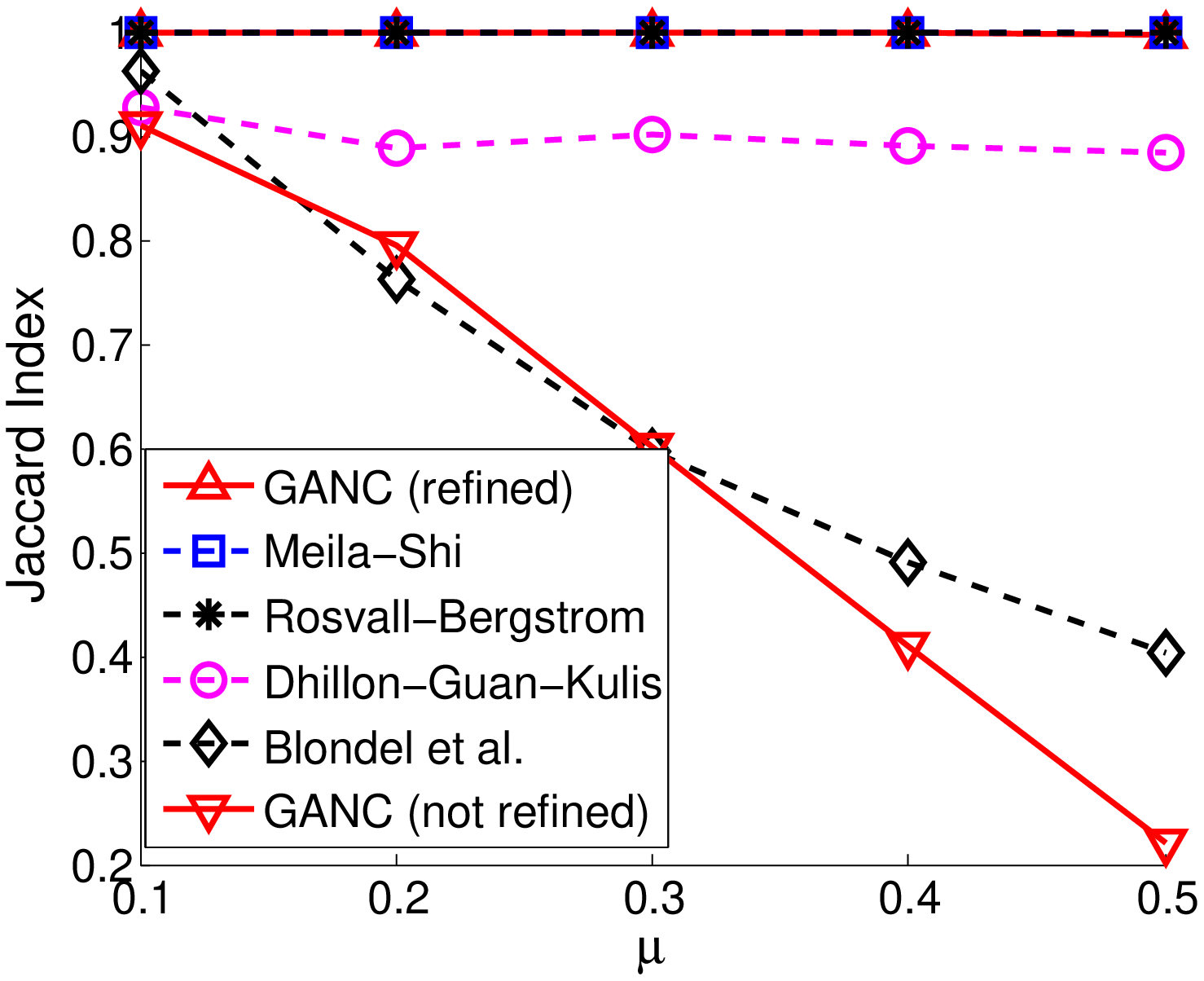} \label{fig:LFR:fig:LFR_5000_ji}} 
\caption{LFR benchmarks with $c_{min}=20, c_{max}=50, d_{max}=30, d_{avg}=25$, and variable proportions of inter-cluster edges ($\mu$). In all of the above figures, Meila-Shi, Rosvall-Bergstrom, and GANC(refined) are overlapping. In (b) and (d) Jaccard index = 1 for the overlapping algorithms that corresponds to perfect extraction of the clusters.} 
\label{fig:LFR:LFR}
\end{figure}

\subsubsection{Maximizing normalized association}

Figure \ref{fig:LFR:LFR} examines how the algorithms perform with respect to maximizing $\NAssoc$ for 1,000-node LFR graphs. In order to observe how the algorithms perform on graphs with heterogeneous clusters, we let the cluster sizes range from 20 to 50 nodes.\footnote{In the real-life networks that we consider in this paper, the clusters have been observed to be of a limited size, regardless of the network size \citep{lekovec2008natural}; for example the Dunbar number suggests an upper limit of 150 nodes for clusters in a social network \citep{dunbar1998grooming}.}
For each value of $\mu$, 100 graph realizations are generated. The value of $\NAssoc$ is divided by $k$ to obtain a value between 0 and 1 which represents the average $\NAssoc$ per cluster.

The algorithms perform almost identically with the exception of the Dhillon et al.\ algorithm \citep{dhillon07}. The refinement step of GANC results in a significant improvement in the value of $\NAssoc$. However, the local search of Dhillon's algorithm does not improve $\NAssoc$ significantly. Note that the Blondel et al.\ algorithm results in higher values of $\NAssoc$ due to selecting smaller number of clusters than the true ones ($\NAssoc(\mathcal{C}_k)/k$ is a monotonically decreasing function of $k$); a fair comparison between algorithms is possible only when the number of clusters are equal.

\subsubsection{Comparing to the planted partitions}
The advantage of exploring performance on synthetic graphs is that a ground-truth partitioning is available. Because the LFR benchmarks are based on the planted partition model, there is the additional advantage that $\NAssoc$ is an appropriate criterion to adopt. Apart from some possible small errors due to the randomness inherent in the construction of the benchmark graphs, the ground truth partitioning will correspond to a maximum of $\NAssoc$ for a given value of $k$.
 
For comparing two partitions on the same graph, we use the Jaccard index \citep{downton80} which for two partitions, $\mathcal{X}$ and $\mathcal{Y}$, is defined as
\begin{align}
  JI = \frac{a}{(a+b+c)}, \label{eq:jaccard_index_external}
\end{align}
where $a$, $b$, and $c$ are, respectively, the total pair of nodes that are assigned to the same cluster in both $\mathcal{X}$ and $\mathcal{Y}$, the same cluster in $\mathcal{X}$ and different clusters in $\mathcal{Y}$, and the same cluster in $\mathcal{Y}$ and different clusters in $\mathcal{X}$. When two partitions are identical, the Jaccard index evaluates to one, and it decreases as the partitions deviate from each other. 

Figures \ref{fig:LFR:fig:LFR_5000_ji} show the closeness of the partitionings identified by the algorithms and the ground truth in terms of the Jaccard index. The algorithm by Blondel et al.\ deviates dramatically from the true clustering (Figure \ref{fig:LFR:fig:LFR_5000_ji}). The reason is that the minimum and maximum cluster sizes are fixed for all the networks in addition to the average node degree. Hence the average number of edges per cluster remains the same making the proportion of the intra-cluster edges decrease. 
Due to the resolution limit of modularity, as the proportion of intra-cluster edges is decreased, modularity maximization algorithms tend to group several clusters into a single cluster. If we reduce $n$ to $1000$, the Blondel et al.\ algorithm leads to closer results to the ground truth.
The algorithm by Dhillon et al.\ does not perform as well as other algorithms that strive to maximize $\NAssoc$ and the local search results in negligible improvement in terms of the Jaccard index.

\begin{figure}[ht] 
\centering 
\subfloat[Ring graph] 	{\hspace*{-.2in} \includegraphics[width=.45\linewidth]{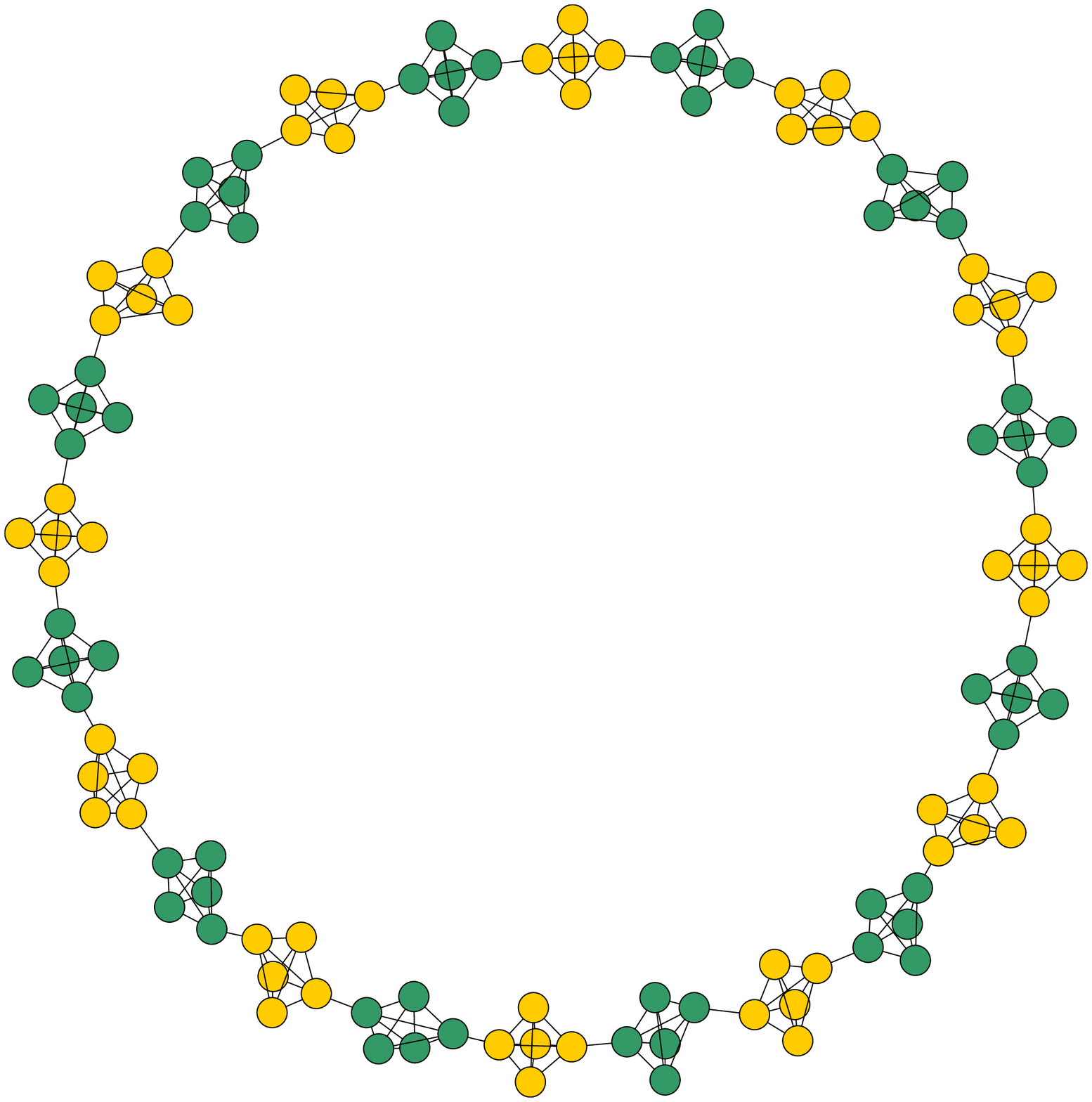} \label{fig:ring:ring_graph}} 
\subfloat[Curvature plot of the ring] {\hspace*{-.in} \includegraphics[width=.6\linewidth]{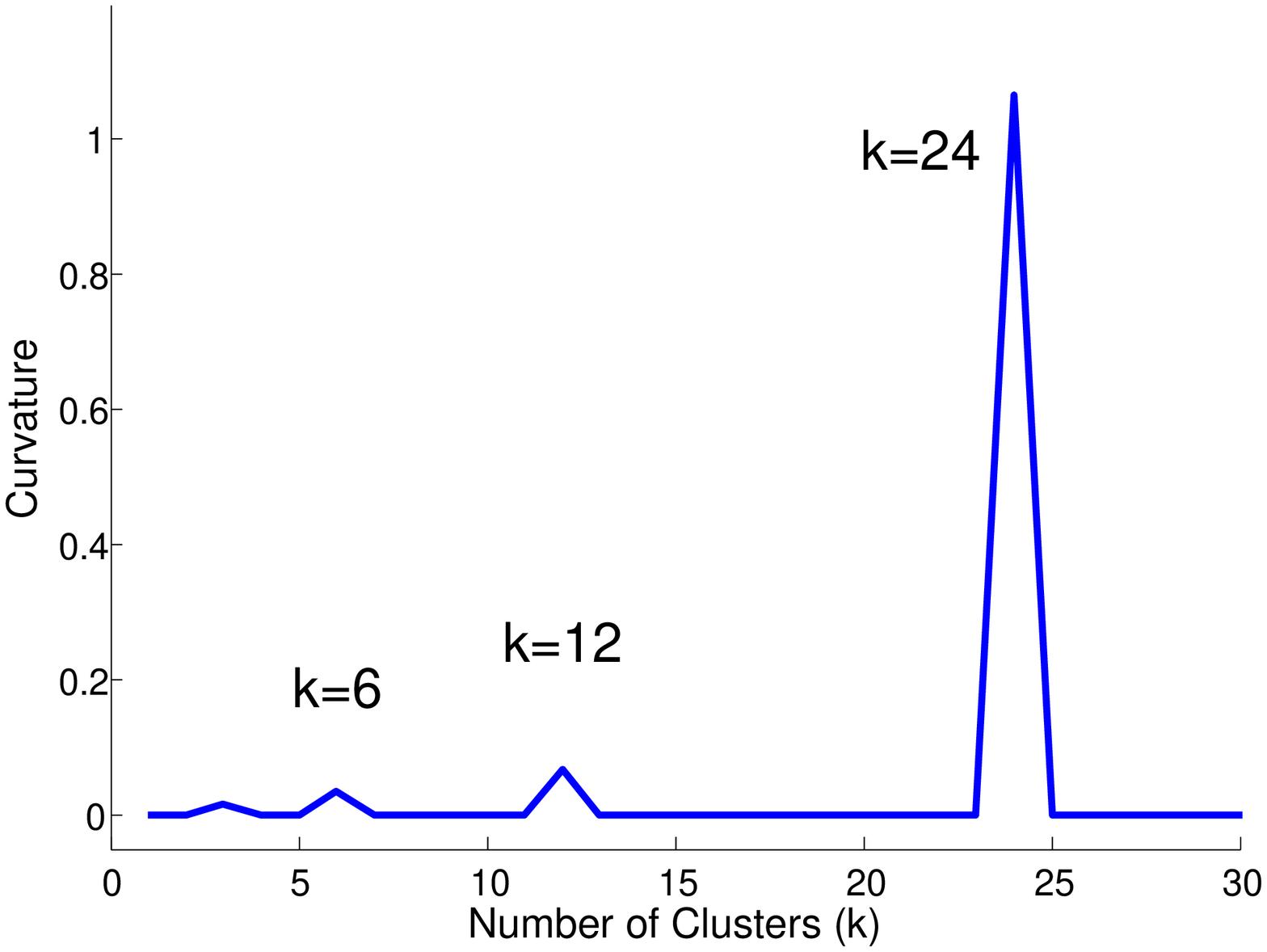} \label{fig:ring:ring_curv}} 
\caption[The ring of 24 cliques.]{The ring of 24 cliques; the highest peak of curvature is at 24.} 
\label{fig:data-pages} 
\end{figure}

\begin{figure}[ht]
  \centering
  \hspace*{-.3in} \includegraphics[width=1.1\linewidth]{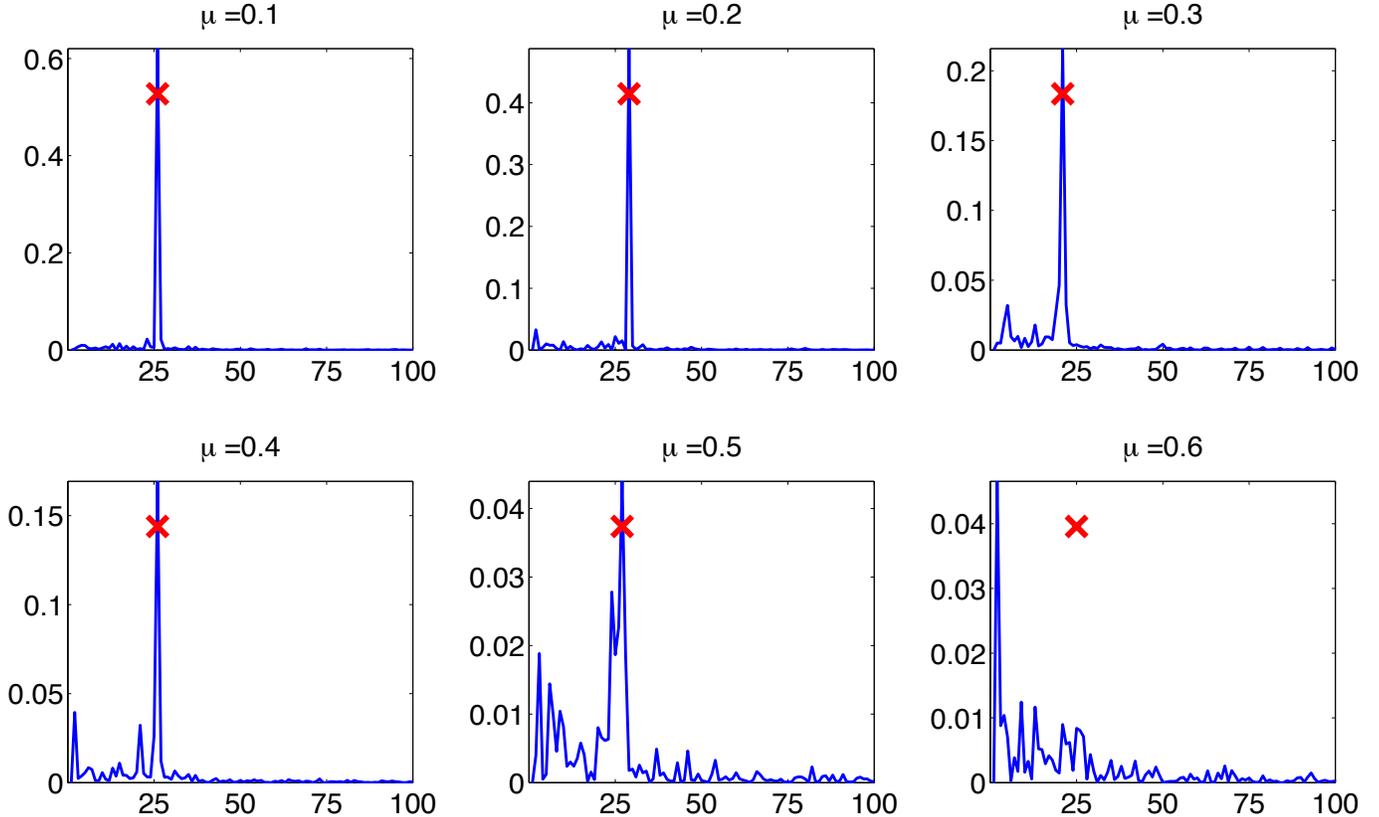}
  \caption[Curvature plot for varying proportion of inter-cluster edges ($\mu$).]{Curvature versus number of clusters for varying proportion of inter-cluster edges ($\mu$). The true clustering indicated by red X. The peak of the curvature becomes less and less distinguishable by increasing $\mu$. LFR benchmark graph with 1000 nodes and $c_{min}=20, c_{max}=50, d_{max}=30, d_{avg}=25$.}  
  \label{fig:LFR:LFR_curv}
\end{figure}

\subsection{Model order selection: the curvature metric}

We first provide an illustrative example of the use of the curvature metric for selecting the number of clusters in the partitioning. This example highlights the difference in behavior compared to the modularity metric used by modularity maximization algorithms~\citep{blondel08}.

An example used by Good et al.\ \citep{good09}, to illustrate the resolution limit is the ring of 24 cliques\footnote{A clique is a subset of nodes in a graph which are fully connected.}, each of which has 5 nodes and is connected to its neighboring clique by a single edge. The value of modularity is maximized by a 12-cluster partition that merges pairs of cliques together. A more natural clustering is to consider each clique as an individual cluster. Figure \ref{fig:ring:ring_curv} shows the curvature plot for the case of the ring of cliques. In addition to the peak of the curvature that is correctly located at 24 clusters, it is interesting to note that the other local peaks of the curvature are also meaningful; e.g., the peak at 12 clusters corresponds to the clustering that maximizes modularity. 

To see how the curvature indicates the true number of clusters for networks with more complex structures, we consider the LFR benchmarks. We fix the LFR parameters for a 1,000-node graph and explore how the curvature changes as $\mu$ is varied. Figure \ref{fig:LFR:LFR_curv} indicates that the peak of the curvature plot correctly identifies the true number of clusters for $\mu$ in the range 0.1 -- 0.5. As $\mu$ increases, the main peak becomes less distinct; for $\mu=0.5$, the second highest peak is almost as large as the primary peak. As the clusters become increasingly interconnected, the curvature plot provides a less clear indication of the true number of clusters. 

Note that Figure \ref{fig:LFR:LFR_curv} corresponds to a single realization of the LFR benchmarks. The empirical probabilities that the highest peak of the curvature corresponds to the true number of clusters for 100 realizations of the LFR benchmark with parameters as in Figure \ref{fig:LFR:LFR_curv} are $p_1=[1.0~1.0~0.96~0.60~0.42~0.06]$, for $\mu = [0.1~0.2~0.3~0.4~0.5~0.6]$ when no refinement step is applied; furthermore the empirical probabilities that one of the two highest peaks corresponds to the true number of clusters are $p_2=[1.0~1.0~1.0~0.98~0.75~0.13]$. 

These probabilities are obtained using the approximate curvatures derived from the agglomerative clustering procedure. To achieve a better insight into the model selection capabilities of the curvature metric, we derive better approximations by applying the refinement step to every level of the hierarchy and then recomputing curvature estimates. With this new procedure, we obtain $p_1 = [1.0~0.99~0.99~0.95~0.87~0.77]$ and $p_2 = [1.0~0.99~0.99~0.96~0.88~0.78]$. This indicates that curvature can provide a good indication of model order even for $\mu=0.6$.

The model order selection algorithm of Rosvall-Bergstrom algorithm quite successfully indicates the true number of clusters in the case of LFR benchmarks; the empirical probabilities for the Rosvall-Bergstrom algorithm is $p_1 = [1.0~1.0~1.0~0.98~0.85~0.83]$. The Blondel et al.\ algorithm does not successfully identify the true number of clusters when $\mu$ increases: $p_1=[1.0~0.99~0.98~0.79~0.07~0.0]$. Note that when $n=5,000$, the failure of the Blondel et al.\ algorithm becomes more evident and $p_1$ turns to a $0\%$ success for any $\mu$ (this is reflected in Figure \ref{fig:LFR:LFR}). However the refined GANC and Rosvall-Bergstrom algorithms obtain the same success ratios when $n=5,000$. 

\subsection{Real networks}

In this section, we examine the behavior the GANC and compare it to the other algorithms for graphs representing real networks. We first examine the behavior for small networks where there is knowledge of the ground truth partition. We then experiment with large networks, which allows us to assess the scalability of GANC.

\subsubsection{Small networks}
We conduct experiments using the Zachary karate club network \citep{zachary}, the football network \citep{girvan02}, and the political books network\footnote{Collected by V. Krebs, \url{http://www.orgnet.com}}. The Zachary karate club network portrays friendships in a karate club. During the polling period, there was a dispute between the manager and the instructor which led to the establishment of a new club by the instructor. The karate students then split into two groups, either staying with the original club or following the instructor to the new club. The two clusters associated with the network correspond to these two groups. Each node in the football networks corresponds to a team in US college football. There are 11 conferences and 5 independent teams. Each edge corresponds to the existence of a match between the connected nodes. The independent teams can be interpreted as outliers. Nodes of the political book network correspond to the political books sold by Amazon.com and are categorized as neutral, conservative, and liberal. The edges between pairs of nodes correspond to frequent co-buying of the pairs of books by the same buyers. The three networks are unweighted. 

Tables \ref{tab:small_real_nets_1} and \ref{tab:small_real_nets_2} compare the performance of GANC with other clustering algorithms for these small, well-known datasets. The tables provide the average $\NAssoc$ per cluster ($\NAssoc/k$) and the Jaccard index. The ``true" number of clusters is indicated within parentheses beside the name of each network\footnote{Note that the ``true" number of clusters is something of an artificial social construct and does not necessarily correspond to a partitioning that maximizes any meaningful graph clustering metric.}. In Table \ref{tab:small_real_nets_1}, all algorithms (including GANC) are informed of the true number of clusters and instructed to generate a partitioning with that number of clusters. Table \ref{tab:small_real_nets_1} indicates that the spectral clustering algorithms (Ng-Jordan-Weiss, Meila-Shi, and Shi-Malik) achieve similar performance in terms of $\NAssoc$. We note that when $k$ is specified for these three networks, GANC identifies partitionings with average $\NAssoc$ as large or larger than those of the partitionings identified by the spectral clustering techniques, with the exception of Shi-Malik in the case of the football network. 
\begin{table}[h]
\renewcommand{\arraystretch}{1.1}
\caption[Small real networks: $k$ known in advance]{Small real networks: $k$ known in advance (average $\NAssoc$ per cluster/Jaccard index). Maximum $\NAssoc$ is indicated by bold text.}
\label{tab:small_real_nets_1}
\centering
{\footnotesize
\begin{tabular}{|c|c|c|c|}
\hline
  & Karate($k=2$) & Football($k=11$) & PolBooks($k=3$) \\
\hline
  GANC (refined)& \textbf{0.872}/0.89 & 0.704/0.820 & \textbf{0.88}/0.67 \\
\hline
  Dhillon-Guan-Kulis & 0.820/0.64 & 0.700/0.820 & 0.87/0.64 \\
\hline
  Ng-Jordan-Weiss & \textbf{0.872}/1.00 & 0.701/0.825 & 0.82/0.57 \\
\hline
  Meila-Shi & 0.869/0.89 & 0.696/0.820 & \textbf{0.88}/0.67 \\
\hline
  Shi-Malik & \textbf{0.872}/1.00 & \textbf{0.706}/0.825 & 0.86/0.67 \\
\hline
\end{tabular}
}
\end{table}

Table \ref{tab:small_real_nets_2} compares the performance of algorithms that select the number of clusters based on some aspect of the data. The number of clusters selected by each algorithm is shown within parentheses after the Jaccard index. Rosvall's algorithm uses a description length metric to select the number of clusters~\citep{rosvall08}; Blondel's algorithm chooses the partitioning that maximizes the modularity \citep{blondel08}. A direct comparison of $\NAssoc$ is not valid when $k$ is not fixed.
\begin{table}[h]
\renewcommand{\arraystretch}{1.1}
\caption{Small real networks: Jaccard index for the estimated model orders.}
\label{tab:small_real_nets_2}
\centering
{\footnotesize
\begin{tabular}{|c|c|c|c|}
\hline
  & Karate & Football & PolBooks \\
\hline
  GANC (curv)& \textbf{0.80} ($k=3$) & 0.76 ($k=13$)  & \textbf{0.69} ($k=2$)) \\
\hline
  Blondel et al.\ & 0.53 ($k=4$) & 0.70 ($k=11$) & 0.67 ($k=3$) \\
\hline
  Rosvall-Bergstrom & 0.71 ($k=3$) & \textbf{0.83} ($k=12$) & 0.63 ($k=5$) \\
\hline
\end{tabular}
}
\end{table}
The first row of this table shows the performance of GANC when the curvature plot is used to select the number of clusters. Although the selected number of clusters does not correspond to the true number of clusters for any of the networks, all three values can be explained. In the karate club network there is a group of students who have weak ties to other members of the network and these are identified as a third cluster; in the football network, GANC isolates the independent teams; in the political books network, GANC does not identify a neutral cluster and assigns each of the neutral books to either the liberal or conservative cluster. 

\subsubsection{Cortical Networks}
Here we study the networks presented in \citep{hagmann08} which are weighted graphs, each with 998 nodes. The nodes correspond to small regions of the human cerebral cortex and the edges correspond to cortico-cortical axonal pathways. The networks are developed for five patients (the extraction is performed twice for patient A). We first perform a comparison of the competing algorithms, then we discuss the clustering results of GANC. 

When applying the Rosvall-Bergstrom and Blondel et al.\ algorithms the used does not have the freedom to choose the number of clusters. Hence we repeat the experiment twice to make a meaningful objective comparison in terms of $\NAssoc$. Table \ref{tab:cortical_nassoc} lists the average $\NAssoc$ of every clustering algorithm and patient. In the first subtable, $k$ is set to the model order selected by the Rosvall-Bergstrom algorithm. Then, in the second subtable, $k$ is set to the model order selected by the Blondel et al.\ algorithm, and in the third subtable, peaks of the model order selected by curvature are used. Note that the Shi-Malik algorithm outperforms the other two spectral algorithms for these datasets; hence we only include the Shi-Malik results.

\begin{table}[h]
\renewcommand{\arraystretch}{1.1}
\caption[]{$\NAssoc$ of the cortical networks. Maximum $\NAssoc$ is indicated by bold text. The model order selected for each column in each sub-table is identified within parentheses.}
\label{tab:cortical_nassoc}
\centering
\begin{adjustwidth}{-.0in}{-.0in}
{\footnotesize
\begin{tabular}{|c|c|c|c|c|c|c|}
\hline
  & A-1 & A-2 & B & C & D & E \\
\hline
  Rosvall-Bergstrom & 0.725 (62) & 0.706 (68) & 0.738 (61) & 0.706 (62) & 0.691 (63) & 0.706 (58) \\
\hline
  GANC (refined) & \textbf{0.730} & \textbf{0.716} & \textbf{0.740} & \textbf{0.712} & \textbf{0.705} & \textbf{0.718} \\
\hline
  Dhillon-Guan-Kulis & 0.696 & 0.682 & 0.712 & 0.688 & 0.670 & 0.684 \\
\hline
  Shi-Malik & 0.715 & 0.692 & 0.727 & 0.690 & 0.686 & 0.694 \\
\hline
\hline
\hline
  Blondel et al.\ & 0.873 (14) & 0.872 (13) & \textbf{0.885} (15) & 0.862 (14) & 0.856 (14) & 0.828 (21) \\
\hline
  GANC (refined) & \textbf{0.886} & \textbf{0.879} & 0.883 & \textbf{0.873} & 0.864 & \textbf{0.840} \\
\hline
  Dhillon-Guan-Kulis & 0.860 & 0.865 & 0.872 & 0.859 & 0.851 & 0.810 \\
\hline
  Shi-Malik & 0.877 & 0.874 & 0.877 & 0.860 & \textbf{0.870} & 0.826 \\
\hline
\hline
\hline
  GANC (refined) & \textbf{0.908} (10) & 0.945 (4) & \textbf{0.899} (12) & \textbf{0.880} (13) & 0.930 (5)& \textbf{0.907}(9) \\
\hline
  Dhillon-Guan-Kulis & 0.889 & 0.946 & 0.897 & 0.863 & 0.916 & 0.885 \\
\hline
  Shi-Malik & 0.898 & \textbf{0.951} & 0.886 & 0.863 & \textbf{0.936} & 0.902 \\
\hline
\end{tabular}
}

\end{adjustwidth}
\end{table}


The clustering results corresponding to the curvature peaks include cluster(s) that contain nodes from both of the brain hemispheres and clusters that include nodes from only one of the hemispheres. In the following discussion, we call the clusters that contain nodes from both of the hemispheres, the central clusters. Graphs A1, B, D, and E include only one central cluster. Graphs A2 and C each include two central clusters. 

The regions of the brain that are commonly grouped in the central clusters are posterior cingulate cortex, precuneus, cuneus, paracentral lobule, pericalcarine cortex, caudal anterior cingulate cortex, isthmuscingulate, isthmus of the cingulate cortex, and lingual gyrus, provided that graph B is excluded. If the second largest peak of the curvature is selected for B ($k=5$), the same regions are assigned to its central cluster. The first five of the mentioned regions are also classified as part of the structural core proposed by Hagmann et al.\ \citep{hagmann08};
Hagmann et al.\ used cluster strengths, cluster degrees, k-Core, s-Core, betweenness centrality, and efficiency\footnote{Cluster strength and degree are the weighted and unweighted degrees, respectively. k-Core (s-Core) is the largest subgraph that includes nodes of degree (respectively, strength) at least k (respectively, s). Betweenness centrality of a region is a measure of the proportion of the shortest paths passing through it. Efficiency of a region indicates how short the region's average path lengths to other regions are.} to propose a structural core of the brain consisting of eight regions.  

The authors of \citep{hagmann08} also used modularity maximization and found six modules, two of which included nodes from both of the hemispheres (central clusters). The regions that are assigned to the central clusters by modularity maximization are similar to the ones extracted by GANC.

%
%
%
%

\subsubsection{Larger networks}

Here we illustrate the performance of our algorithm on large graphs. 
We apply our algorithm to four networks with different natures: Cond-Mat (a collaboration network) \citep{newman01}, Googleweb (a web graph) \citep{lekovec2008natural}\footnote{Google programming contest: \url{http://www.google.com/programming-contest/}}, Amazon (a product co-purchasing network)\citep{leskovec2007dynamics}, and AS-Skitter (an autonomous system graph)\citep{leskovec2005graphs}. The nodes in Cond-Mat represent scientists that submit their preprints to the condensed matter archive at \url{www.arxiv.org}; the edges represent co-authorships. The nodes in Googleweb represent websites and the directed edges represent the existence of hyperlinks. Each node in Amazon network corresponds to a product purchased at \url{www.amazon.com}. Each directed edge from a node to another means that when the former is purchased, the latter is frequently also purchased. AS-Skitter is a network of autonomous systems extracted by traceroute analysis.\footnote{\url{http://www.caida.org/tools/measurement/skitter}} 

The Cond-Mat graph is weighted and the rest of the graphs are unweighted. In Table \ref{tab:large_real_nets}, the average and maximum degrees do not take the weights into account; the number of edges affects the run-time of GANC, not the weight values. 
None of the above graphs are originally connected. However each has a very large connected component that includes the majority of the nodes and the edges. Here we conduct clustering analysis of the largest connected component. 
Some of these graphs are directed, but we construct undirected graphs by adding the adjacency matrix to its transpose.


\begin{table*}[t]
\renewcommand{\arraystretch}{1.1}
\thispagestyle{empty}\caption{Larger real networks: Partition metrics ( (Number of clusters) $\NAssoc$ per cluster/execution time).}
\label{tab:large_real_nets}
\centering
\begin{adjustwidth}{-.75in}{-.0in}
{ \small \begin{tabular}{|c||c|c|c|c|}
\hline
  Network & Cond-Mat & Googleweb & Amazon  & AS-Skitter \\
\hline & \\[-1.45em]\hline
  \# of Nodes/Edges & 36458/171736 & 342408/1142134 & 403364/2443311  & 1694616/11094209 \\
\hline
  Avg/Max Degree & 28.5/278 & 6.7/1367 & 12.1/2752  & 13.1/35455 \\
\hline & \\[-1.45em]\hline
  GANC & (2121) \textbf{0.784}/\textbf{0m2s} & (12213) \textbf{0.869}/2m30s & (11237) \textbf{0.768}/\textbf{3m4s} & (26399) \textbf{0.769}/\textbf{46m33s} \\
\hline
  Dhillon-Guan-Kulis & (2121) 0.669/0m3s & (12213) --- & (11237) 0.679/6m0s & (26399) --- \\
\hline
  Rosvall-Bergstrom & (2121) 0.746/0m10s & (12213) 0.844/\textbf{1m17s} & (11237) 0.712/11m26s & (26399) 0.720/99m37s \\
\hline & \\[-1.45em]\hline
  GANC & (82) \textbf{0.957}/0m2s & (239) \textbf{0.993}/2m30s & (230) \textbf{0.984}/3m4s &  (1776) \textbf{0.933}/46m33s \\
\hline
  Dhillon-Guan-Kulis & (82) 0.784/\textbf{$<$1s} & (239) 0.935/\textbf{0m2s} & (230) 0.872/\textbf{0m4s} &  (1776) 0.709/8m0s \\
\hline
  Blondel et al.& (82) 0.849/0m1s & (239) 0.990/0m5s & (230) 0.952/0m14s &  (1776) 0.859/\textbf{0m4s} \\
\hline & \\[-1.45em]\hline
  GANC & (37) 0.970/0m2s & (7) 0.999/2m30s & (22) 0.997/3m4s & (32) 0.994/46m33s \\
\hline
  GANC & (526) 0.902/0m2s & (855) 0.984/2m30s & (206) 0.996/3m4s & --- \\
\hline
\end{tabular} }
\end{adjustwidth}
\end{table*}


Table \ref{tab:large_real_nets} compares the performance of the algorithms that are scalable to such large networks. The table shows the average $\NAssoc$ per cluster of the identified partitioning and the time required for completion of the algorithm. The spectral clustering algorithms cannot be executed on our test machine when either the number of nodes or the number of clusters is very large. Both the computational time and the memory requirements are excessive. We therefore compare the performance of GANC, Dhillon-Guan-Kulis (no local search), Rosvall-Bergstrom, and the Blondel et al.\ algorithms. 

The Rosvall-Bergstrom and Blondel et al.\ algorithms automatically select the number of clusters. A meaningful comparison of the average $\NAssoc$ is only possible when the number of clusters are equal; to facilitate comparisons, we therefore construct multiple partitionings for GANC and the Dhillon-Guan-Kulis algorithms, each with a different number of clusters. We also construct a partitioning corresponding to one or more of the peaks in the curvature plot (for some of the networks, there are two peaks that are very similar in value, so we consider it useful to examine the partitionings corresponding to each). The last two rows of Table \ref{tab:large_real_nets} correspond to those peaks. 

GANC is superior to other algorithms with respect to maximizing $\NAssoc$ in all of the cases. GANC also significantly outperforms the Dhillon-Guan-Kulis algorithm, which is also striving to maximize $\NAssoc$. The latter is also outperformed by Rosvall-Bergstrom and Blondel et al.\ algorithms. 

In terms of the computation time, GANC is often faster than Rosvall-Bergstrom, but the scaling behavior is different. To illustrate this, in addition to the graphs listed in Table \ref{tab:small_real_nets_1}, we have applied the algorithms to the road network graph of California \citep{lekovec2008natural} which is extremely sparse (the average degree is 2.8). The graph contains around 2 million nodes. While GANC performs the clustering in 32 seconds, in takes 197 minutes for the Rosvall-Bergstrom algorithm to converge to a solution.
The complexity of GANC is dominated by the agglomerative clustering procedure which requires $O(mh\log n)$ operations. Hence the speed of GANC depends on the number of edges in the graph, and the height of the dendrogram.
However the graph sparsity does not show an impact on Rosvall-Bergstrom's execution time. Blondel's algorithm is much faster than GANC, but as mentioned previously, it suffers from resolution limit associated with clustering algorithms that maximize modularity. 
The Dhillon-Guan-Kulis algorithm is also very fast for small values of $k$, but it gets slower and its memory requirements become excessive if $k$ becomes large. 

\begin{figure}[ht]
\centering 
\subfloat[Amazon] {\hspace*{-.4in} \includegraphics[width=.55\linewidth]{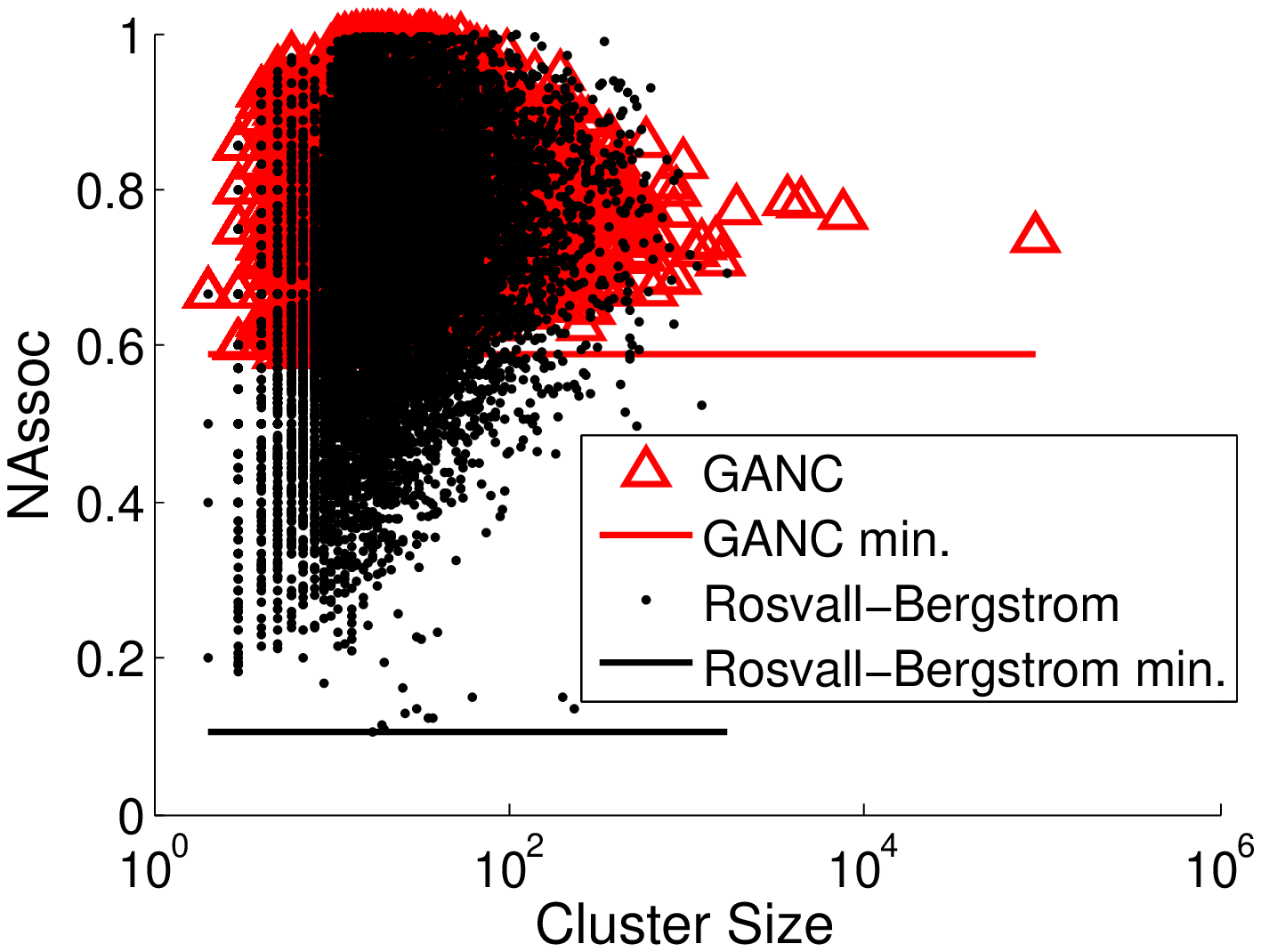} \label{fig:individual_nassoc_amazon}}
\subfloat[US Patent Citation] {\hspace*{-.1in} \includegraphics[width=.55\linewidth]{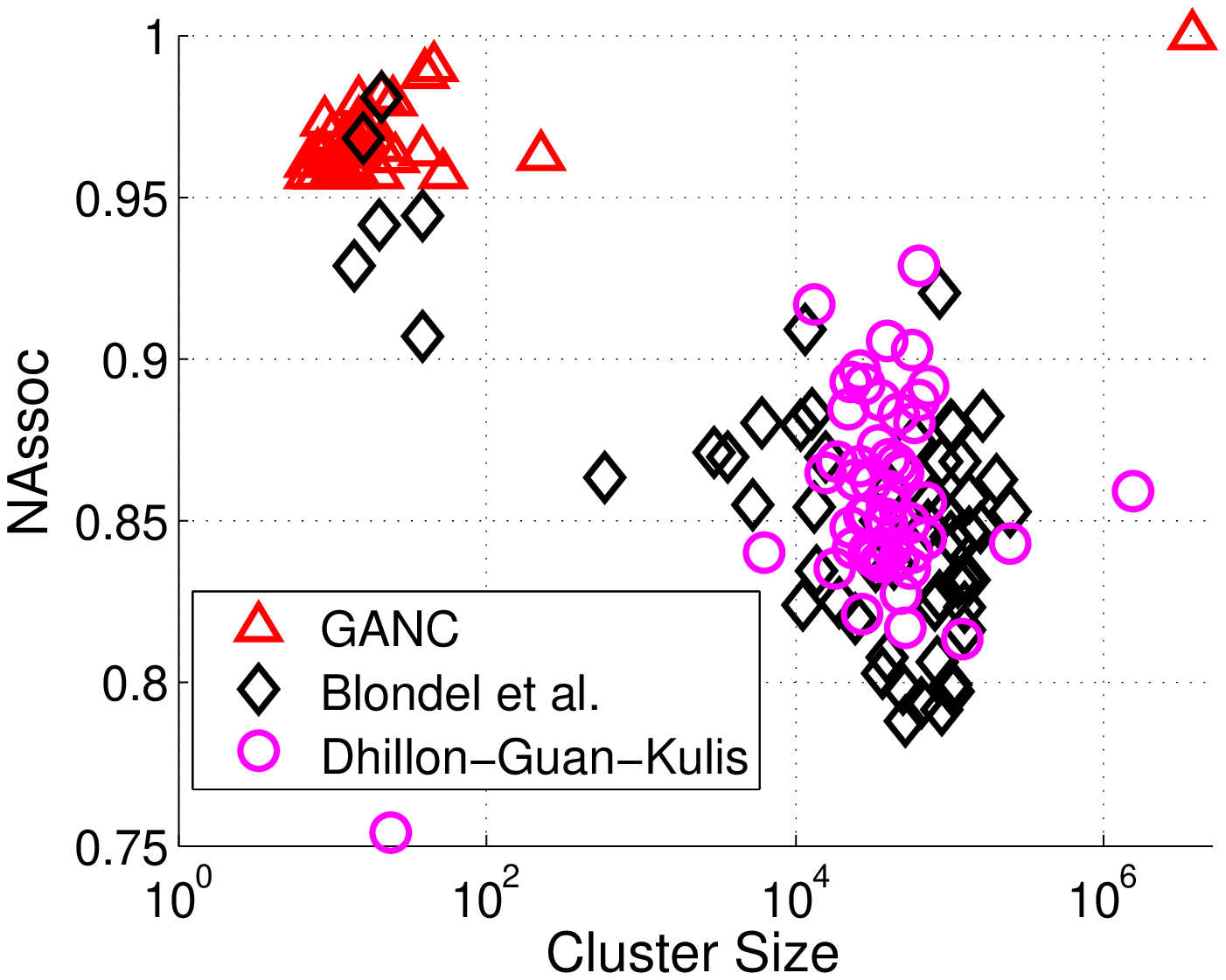} \label{fig:individual_nassoc_citation}} 
\caption{$\NAssoc$ of individual clusters.} 
\label{fig:individual_nassoc} 
\end{figure}

To illustrate the properties of the algorithm outputs in terms of the cluster sizes and the connectivity of individual clusters, we have examined the plots of $\NAssoc$ of individual clusters versus the number of nodes in them. Each algorithm behaves similarly on different graphs presented in Table \ref{tab:large_real_nets}. Figure \ref{fig:individual_nassoc_amazon} compares GANC and Rosvall-Bergstrom when applying these algorithms on Amazon co-purchasing graph. For the same number of clusters, GANC produces clusters with higher values of $\NAssoc$ than Rosvall-Bergstrom. The latter produces many small clusters with very low values of $\NAssoc$. 
The behavior of the Dhillon-Guan-Kulis and the Blondel et al.\ algorithms are discussed in our case study. 

\subsection{Case Study: US Patent Citation Graph}
As a case study, we consider the undirected version of the citation graph released by the National Bureau of Economic Research \citep{website:patents, hall01}. The patents are classified into 6 broad technological categories. A more refined classification leads to 36 sub-categories. We use each patent's label (category or sub-category) in addition to $\NAssoc$ and runtime to perform a comparison. 

The original graph is not connected; however the largest connected component contains more than 99.7\% of the nodes and all of the edges. Hence we focus on the largest connected component which contains 3,764,117 nodes and 16,511,740 edges. The maximum node degree is 793. 
\subsubsection{Clustering Runtime}
The Rosvall-Bergstrom algorithm \citep{rosvall08} was terminated after more than 30 hours without converging to a solution. GANC takes 77 minutes to construct the full hierarchy. After the hierarchy has been generated, each flat partitioning including the refinement takes less than 35 seconds. The algorithm by Blondel et al.\ \citep{blondel08} takes 8 minutes. The Dhillon-Guan-Kulis algorithm \citep{dhillon07} takes 72 seconds for $k=57$ and increases as $k$ is increased. 

\subsubsection{Maximization of Normalized Association}
In order to have a fair comparison in terms of $\NAssoc$, we fix $k=57$ to match the number of clusters of the Blondel et al.\ result. The values of $\NAssoc/k$ for GANC, Dhillon-Guan-Kulis, and Blondel et al.\ are 0.964, 0.859, and 0.855, respectively. The values of $\NAssoc$ of the individual clusters are shown in Figure \ref{fig:individual_nassoc_citation} which illustrates the clear superiority of GANC.
\subsubsection{Extraction of True Clusters and Absence of Large Well-Defined Clusters}
\begin{figure}[ht]
\centering 
\subfloat[Maximum Homogeneity Proportions] {\hspace*{-.4in} \includegraphics[width=.55\linewidth]{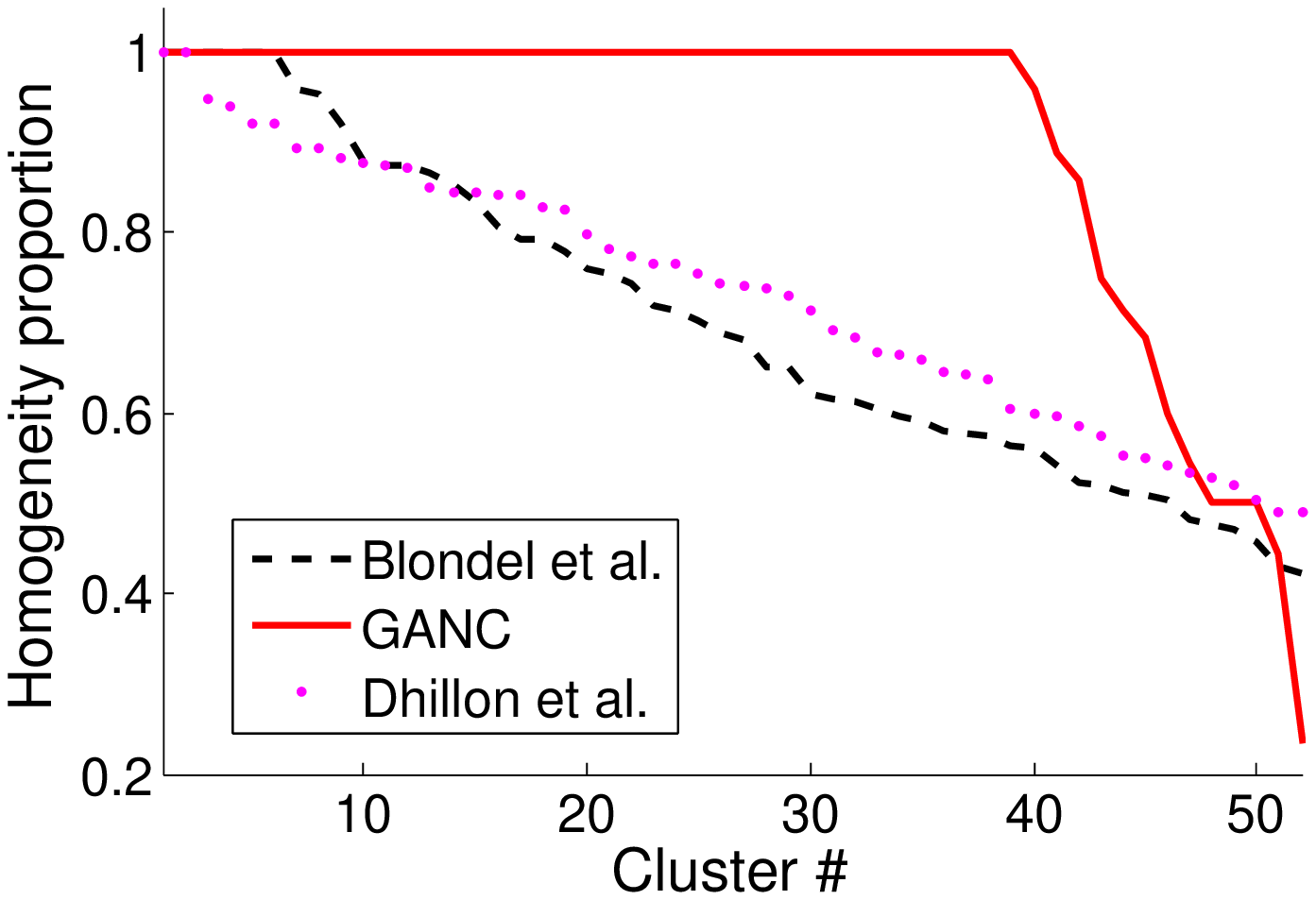} \label{fig:CITATION_1:1}}
\subfloat[Homogeneity Proportion per Cluster] {\hspace*{-.1in} \includegraphics[width=.6\linewidth]{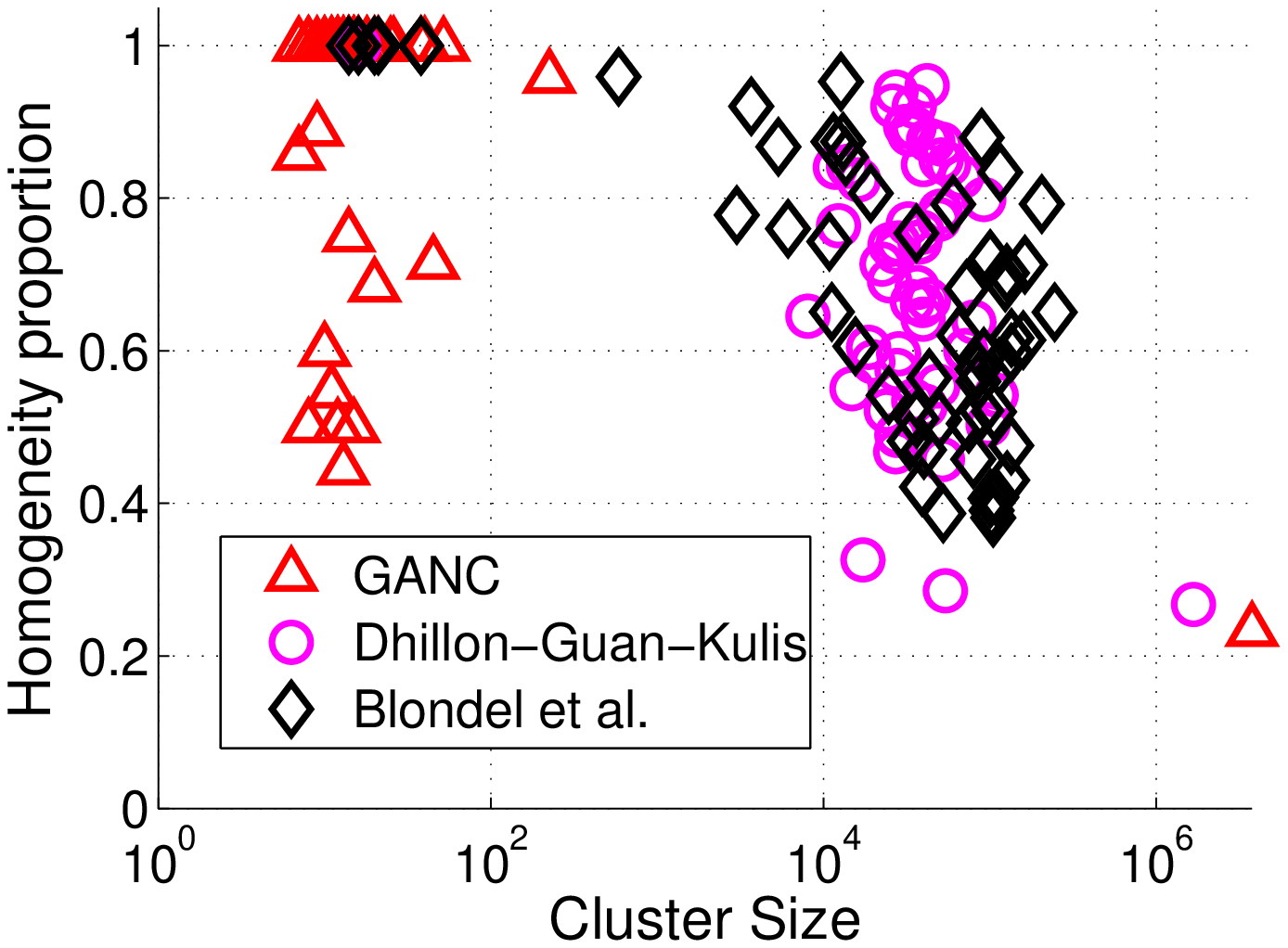} \label{fig:CITATION_2:2}} 
\caption{The 52 most homogeneous clusters out of 57 clusters using the Blondel et al.\ and the Dhillon-Guan-Kulis algorithms.} 
\label{fig:CITATION_1} 
\end{figure}

We use the categories and sub-categories to classify the nodes in the patent citation graph. We denote the maximum proportion of nodes from the same class in a given cluster as the {\em homogeneity proportion} of that cluster. For the Blondel et al.\ and Dhillon-Guan-Kulis algorithms, we use $k=57$ and for GANC we use $k=52$ (the closest peak of the curvature plot). The clusters are sorted according to their homogeneity proportion in Figure \ref{fig:CITATION_1:1}. The figure shows the superiority of GANC in extracting clusters of nodes from the same categories. When sub-categories are employed for the assessment, the superiority of GANC becomes more pronounced.

Figure \ref{fig:CITATION_1:1} alone does not provide a fair comparison as any singleton would have homogeneity proportion of 1. In Figure \ref{fig:CITATION_2:2}, the homogeneity proportion is plotted versus the size of the extracted clusters for the three algorithms. 


The Blondel et al.\ algorithm produces clusters as small as 14 to as large as 250,000 nodes. However the largest homogeneous cluster that is extracted has 38 nodes. The quality of clusters degrade as the size of the clusters increase. This trend is very likely to be due to the resolution limit of modularity.

Both GANC and Dhillon-Guan-Kulis produce a very large cluster (the core cluster \citep{lekovec2008natural}). Excluding the core, Dhillon-Guan-Kulis produces clusters of an average size of $n/k$ which is around 37,000 nodes in this example (See Figure \ref{fig:CITATION_2:2}). This behavior of the Dhillon-Guan-Kulis algorithm is due to the underlying region-growing procedure it adopts from Metis \citep{Karypis99}. 

The clusters extracted by GANC are much smaller than the other two algorithms. Excluding the core, the cluster sizes are between 7 and 225. 
Similar to previous observations of Leskovec et al.\ \citep{lekovec2008natural}, the strongly-connected clusters of the large graphs listed in Table \ref{tab:large_real_nets} diminish in number as we move towards the center of the graph. Having a small homogeneity proportion is then expected for the core as it includes several clusters that are not very well-connected and are lumped together. If we study a clustering that is in a lower level in the hierarchy, further clusters would be extracted from the core and the core shrinks. The resulting clusters are not as well-connected in terms of $\NAssoc$ though. In other words, the most isolated and well-connected clusters are extracted at the higher levels of the hierarchy. 

\section{Conclusion}
\label{sec:conc}
We proposed a novel algorithm to maximize normalized association that consists of three steps: the agglomerative hierarchical clustering procedure, the model order selection step, and the refinement step. 

The agglomerative clustering procedure requires $O(n\log^2 n)$ operations for many real-life graphs, where $n$ is the number of nodes in the graph. This procedure dominates the computational complexity of the other steps of the algorithm. The second step of our algorithm, the model order selection, is based on the relative improvement of normalized association when changing the number of clusters; the {\em curvature} plot is used to select one or several model orders for the final clustering. Unlike modularity, the curvature metric does not exhibit an intrinsic resolution limit. For a multi-resolution analysis, a user can specify a range on the allowable number of clusters, and the algorithm will select the number of clusters with the maximum curvature in that range. After selecting the number of clusters, the clusters are passed to the refinement step. This step of our algorithm iterates over the boundary nodes in the clusters and explores possible improvements in normalized association by moving each of the boundary nodes to their neighboring clusters. Using the map data structure, the overhead added by the refinement becomes negligible. Experiments show that despite the negligible runtime of the refinement step, it significantly improves the initial results with respect to the normalized association maximization. 

Our experimental analysis on relatively small networks indicated that
the proposed algorithm identified partitions that have normalized
association values comparable to the spectral algorithms that involve
an eigendecomposition. These algorithms are too computationally
complex to be applied to very large graphs. We demonstrated that our
proposed algorithm can be applied to large graphs (millions of nodes
and edges). For these large graphs, the proposed algorithm identifies partitions that
have larger values of normalized association than those identified by the
only comparable algorithm that directly addresses the normalized cut
metric.

A clustering algorithm can by no means be suitable for every application. For example, despite the failure of Dhillon-Guan-Kulis algorithm to extract clusters with nodes of the same categories in the patent citation network, it generates clusters of very similar sizes. This is more suitable for VLSI applications for instance \citep{li06}. 
On the other hand, when clustering is meant to extract ``communities'' (group of nodes with strong intra-connection), GANC and Rosvall-Bergstrom are clearly preferred. 

The Blondel et al.\ and Rosvall-Bergstrom algorithms are not able to generate a partitioning to an arbitrary number of clusters\footnote{The available implementation of the Rosvall-Bergstrom algorithm is based on the Blondel et al.\ algorithm. See \url{http://www.tp.umu.se/~rosvall/algorithm.pdf}.}; even though they are agglomerative, they do not generate the complete hierarchy because they merge several nodes/clusters in each of their iterations. 
Hence if one is interested in several arbitrary clustering levels, GANC fits one's requirement the best out of the competing algorithms discussed.  

In this paper we only considered undirected graphs while the edge directions could carry valuable information about the structure of a graph. An extension of GANC can be developed by adopting the {\em generalized normalized cut} criterion~\citep{meila2007clustering}.




\small
\bibliographystyle{elsarticle-num}
\bibliography{GANC_manuscript}







\end{document}